%% file: 0_Main.tex
\documentclass[sigconf]{acmart}
\AtBeginDocument{%
  }

\usepackage{amsfonts}

\usepackage{amssymb}
\usepackage{bm}
\usepackage{nicefrac}

\usepackage{multirow}

\usepackage{subcaption} 

\usepackage{enumitem}   

\usepackage{xspace}

\newtheorem{lemma}{Lemma}
\newtheorem{remark}{Remark}
\newtheorem{assumption}{Assumption}
\usepackage{colortbl}
\usepackage{makecell}
\definecolor{mygray}{gray}{0.9}

\begin{document}

\title{Adaptive Graph Mixture of Residual Experts: Unsupervised Learning on Diverse Graphs with Heterogeneous Specialization}


\author{Yunlong	Chu}
\affiliation{%
  \institution{School of New Media and Communication, Tianjin University}
  \city{Tianjin}
  \country{China}}
\email{2024245030@tju.edu.cn}

\author{Minglai	Shao}
\authornote{Corresponding author.}
\affiliation{%
  \institution{School of New Media and Communication, Tianjin University}
  \city{Tianjin}
  \country{China}}
\email{shaoml@tju.edu.cn}

\author{Zengyi Wo}
\affiliation{%
  \institution{Baidu}
  \city{Beijing}
  \country{China}
}
\email{wozengyi1999@tju.edu.cn}

\author{Bing Hao}
\affiliation{%
  \institution{School of New Media and Communication, Tianjin University}
  \city{Tianjin}
  \country{China}}
\email{haobing@tju.edu.cn}

\author{Yuhang Liu}
\affiliation{%
  \institution{School of New Media and Communication, Tianjin University}
  \city{Tianjin}
  \country{China}
}
\email{liuyuhang_13@tju.edu.cn}

\author{Ruijie Wang}
\authornotemark[1]
\affiliation{%
  \institution{School of Computer Science and Engineering, Beihang University}
  \city{Beijing}
  \country{China}}
\email{ruijiew@buaa.edu.cn}

\author{Jianxin	Li}
\affiliation{%
  \institution{School of Computer Science and Engineering, Beihang University}
  \city{Beijing}
  \country{China}}
\email{lijx@buaa.edu.cn}




\begin{abstract}
Graph Neural Networks (GNNs) face a fundamental adaptability challenge: their fixed message-passing architectures struggle with the immense diversity of real-world graphs, where optimal computational strategies vary by local structure and task. While Mixture-of-Experts (MoE) offers a promising pathway to adaptability, existing graph MoE methods remain constrained by their reliance on supervised signals and instability when training heterogeneous experts. We introduce ADaMoRE (Adaptive Mixture of Residual Experts), a principled framework that enables robust, fully unsupervised training of heterogeneous MoE on graphs. ADaMoRE employs a backbone-residual expert architecture where foundational encoders provide stability while specialized residual experts capture diverse computational patterns. A structurally-aware gating network performs fine-grained node routing. The entire architecture is trained end-to-end using a unified unsupervised objective, which integrates a primary reconstruction task with an information-theoretic diversity regularizer to explicitly enforce functional specialization among the experts. Theoretical analysis confirms our design improves data efficiency and training stability. Extensive evaluation across 16 benchmarks validates ADaMoRE's state-of-the-art performance in unsupervised node classification and few-shot learning, alongside superior generalization, training efficiency, and faster convergence on diverse graphs and tasks.
\end{abstract}





\maketitle

\input{1_Introduction}
\input{2_Preliminary}
\input{3_Method}
\input{4_Experiments}
\input{5_Related_Works}
\input{6_Conclusion}


\bibliographystyle{ACM-Reference-Format}
\bibliography{Reference}

\appendix
\input{7.1_Appendix}

\end{document}

%% file: 1_Introduction.tex
\section{Introduction}
\label{sec:introduction}
Graph Neural Networks (GNNs) have demonstrated remarkable success in modeling graph-structured data across various Web applications, such as search engines, social networks, etc~\cite{gcn, GAT, hamilton2018inductiverepresentationlearninglarge}. Conventional GNNs typically employ a "one-size-fits-all" message-passing mechanism, which performs effectively on graphs with uniform structural properties. However, real-world graphs are rarely uniform; they exhibit immense \textbf{diversity} in terms of connectivity patterns, node attributes, and the nature of downstream tasks. A GNN architecture optimized for one type of structure often fails when applied to another~\cite{Wu_2021, zheng2023findingmissinghalfgraphcomplementary}. This challenge is further exacerbated when considering the variety of downstream tasks that a single graph may need to support, fundamentally limiting the development of robust, adaptive graph learning models.

To overcome the limitations of fixed-architecture GNNs, one line of research focuses on designing \textbf{adaptive GNNs}. These approaches dynamically adjust their computational patterns based on the graph's properties. Strategies range from employing a combination of complementary graph filters~\cite{GREET, S3GCL} to developing learnable polynomial filters~\cite{earnable_polynomial_filters}. However, such specialized designs still require model fine-tuning and often lack the extensibility and generalizability to handle the broader diversity of graph structures and tasks, limiting their potential on larger-scale graph data. A more general and powerful paradigm for adaptivity is the \textbf{Mixture-of-Experts}~\cite{adaptivity, 6796382, CHEN19991229}. The MoE paradigm enables conditional computation by dynamically routing inputs to specialized experts. This technique has proven highly effective for scaling large neural networks, achieving state-of-the-art results in domains such as natural language processing and computer vision~\cite{expert_1, expert_2, expert_3}. This paper investigates the MoE framework for graph-structured data, aiming to create an adaptive solution that accommodates a wide range of graph variations with low customization overhead. 

Despite growing interest in Mixture-of-Experts (MoE) for graph learning~\cite{GMoE, ma2024mixturelinkpredictorsgraphs,li2025hierarchicalmixtureexpertsgeneralizable,wang2025cooperationexpertsfusingheterogeneous}, existing methods face two fundamental limitations that restrict their practical utility: \textbf{Limitation 1: Heavy reliance on supervised signals}. Current approaches are heavily dependent on task-specific labels to train their experts or guide the gating network. This reliance on explicit supervision fundamentally limits their applicability, as it prevents generalization to the vast majority of real-world graphs where labels are scarce or expensive to acquire. Consequently, these models fail to learn intrinsic, transferable graph representations, making them unsuitable for unsupervised or zero-shot scenarios. 

\begin{figure*}[t]
    \centering
    \captionsetup{skip=3pt}
    
    \begin{subfigure}[t]{0.32\textwidth}
        \centering
        \includegraphics[width=\linewidth]{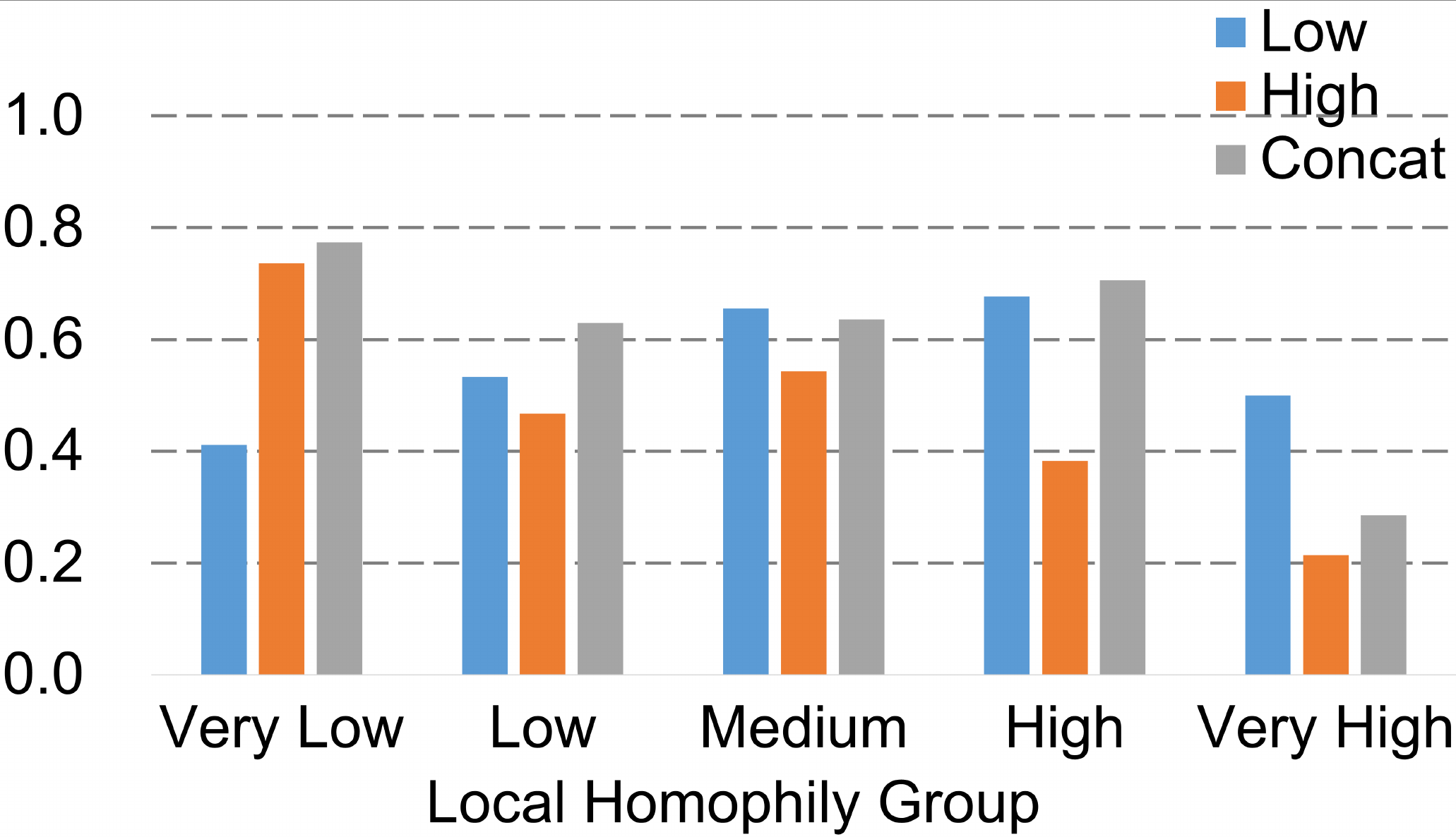}
        \subcaption{Filter vs. Local Homophily.}
        \label{fig:motivation_a}
    \end{subfigure}
    \hfill
    \begin{subfigure}[t]{0.32\textwidth}
        \centering
        \includegraphics[width=\linewidth]{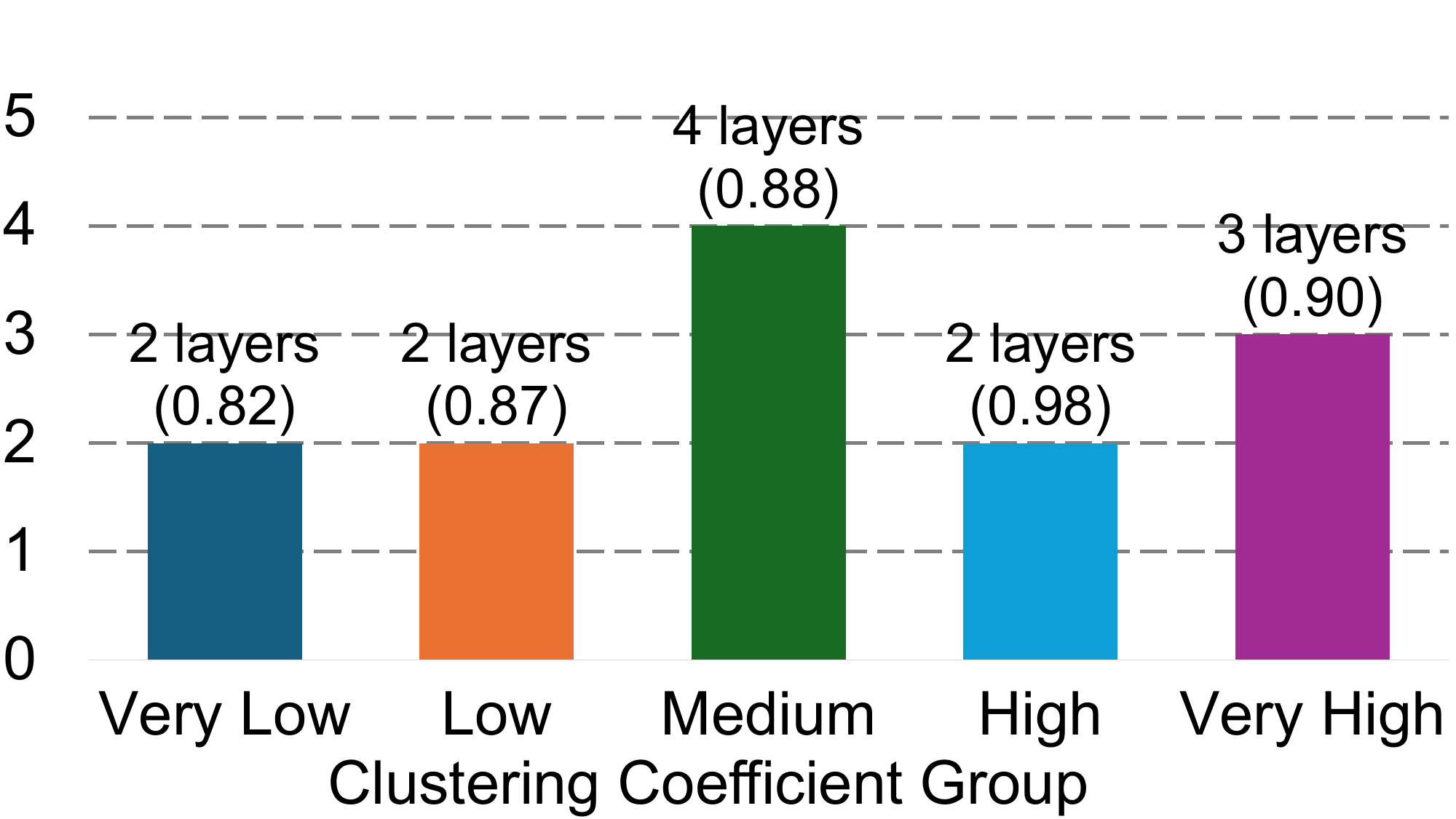}
        \subcaption{Depth vs. Clustering Coefficient.}
        \label{fig:motivation_b}
    \end{subfigure}
    \hfill
    \begin{subfigure}[t]{0.32\textwidth}
        \centering
        \includegraphics[width=\linewidth]{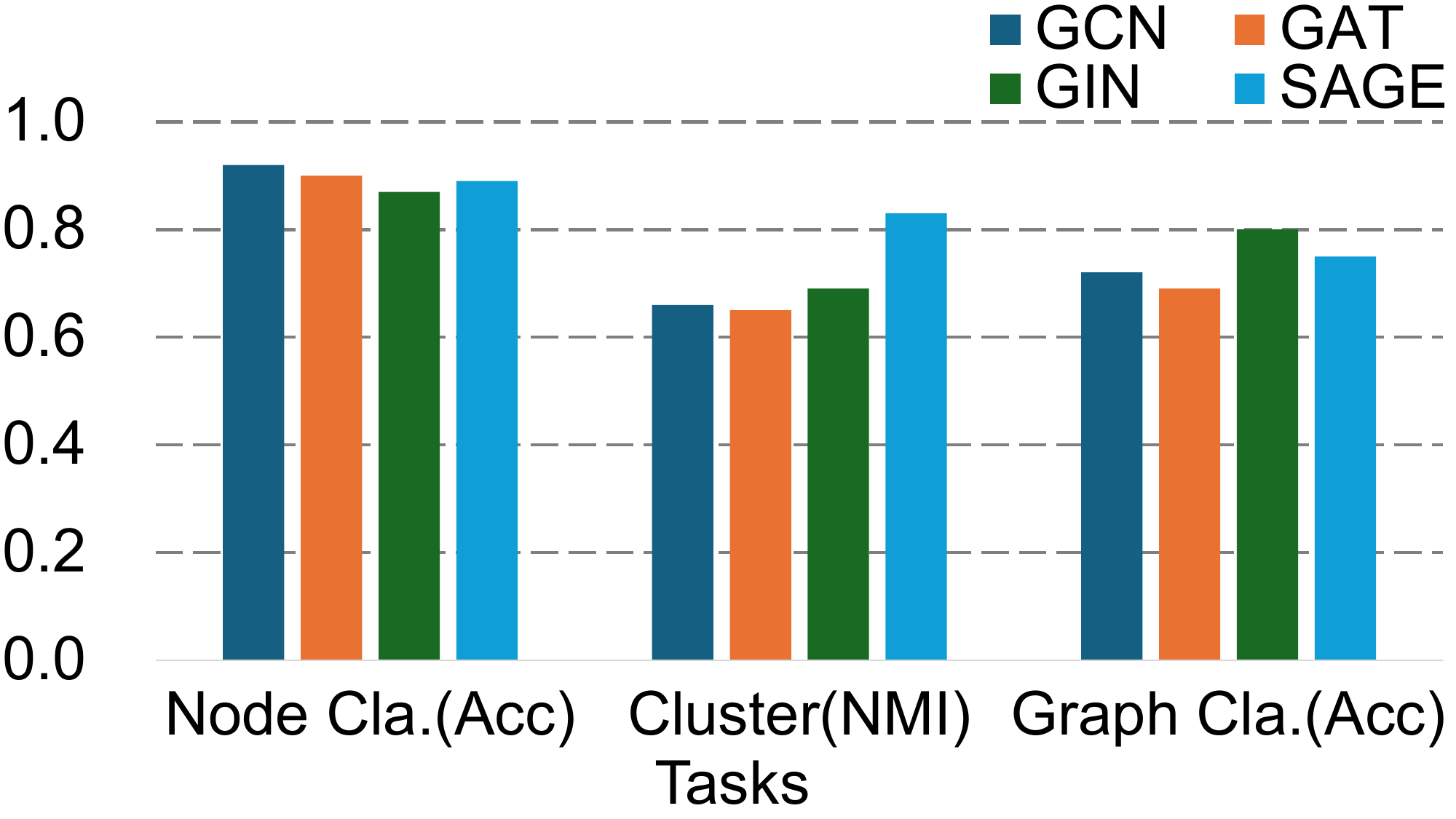}
        \subcaption{Architecture vs. Task.}
        \label{fig:motivation_c}
    \end{subfigure}

    \caption{The "one-size-fits-all" paradigm of GNNs is suboptimal. The ideal GNN configuration is highly context-dependent, varying with (a) local structural context, (b) node topological roles, and (c) the specific downstream task.}
    \label{fig:motivation}

    \Description{A composite figure with three plots demonstrating the need for adaptive GNNs. The plots show that the optimal GNN filter, depth, and architecture vary depending on the node's local structure, its topological role, and the downstream task, respectively.}
\end{figure*}

\textbf{Limitation 2: Instability in training heterogeneous experts}. While most graph MoE models use homogeneous experts to increase capacity, this design is inherently limited. It cannot capture the diverse computational patterns required for complex graph data/tasks, as no single GNN architecture is universally optimal. Our preliminary analysis confirms this: optimal model choices are highly context-dependent. For instance, on the Roman-Empire dataset, the most effective graph filter varies significantly for nodes with different local homophily levels (Figure~\ref{fig:motivation_a}). Similarly, the ideal model depth is not uniform; on the Cora dataset, nodes with low clustering coefficients benefit from shallow architectures, while those in densely-connected regions require deeper models (Figure~\ref{fig:motivation_b}). This architectural preference extends to the task level, where a GNN optimal for node classification on Cora may be suboptimal for graph classification (Figure~\ref{fig:motivation_c}). These observations strongly motivate the use of heterogeneous experts. However, pioneering works in this direction rely on complex expert search or customized training strategies, which our theoretical analysis shows to be data-inefficient and prone to instability, especially in an unsupervised setting.

These two limitations underscore a critical research gap: the absence of a principled and general framework for training a heterogeneous MoE on graphs in a fully unsupervised manner. To bridge this gap, we propose \textbf{ADaMoRE} (Adaptive Mixture of Residual Experts), a novel framework for robust and adaptive graph representation learning. ADaMoRE is built on a backbone-residual architecture, where a stable baseline representation from foundational experts is enhanced by a separate pool of specialized residual experts via residual connection. The framework introduces two key components to enable stable unsupervised training: (1) an information-theoretic diversity regularizer that explicitly promotes functional specialization among experts to mitigate optimization interference, and (2) a structurally-aware gating mechanism that performs fine-grained, node-specific routing based on local topological patterns. Our theoretical analysis demonstrates that this architecture significantly improves data efficiency and enhances training stability during pre-training by closing training errors, thereby directly addressing Limitation 2. To overcome Limitation 1 (dependence on supervision), we employ masked feature reconstruction as the unsupervised pre-training objective. This is complemented by a dedicated cross-filter reconstruction loss designed to train the structurally-aware gating mechanism without any labels. Compared to existing approaches, ADaMoRE's core contribution lies in its principled integration of a heterogeneous expert architecture with a unified unsupervised training paradigm, enabling stable and effective graph representation learning without reliance on labeled data.

In summary, our contributions are as follows:
\begin{itemize}[nosep, leftmargin=*]
    \item We systematically analyze the limitations of fixed GNN architectures on diverse graphs and identify the potential of leveraging a mixture of heterogeneous GNN experts for a more adaptive and generalizable framework.
    \item We propose ADaMoRE, a novel MoE architecture for graphs. Its superiority stems from a robust backbone-residual design that effectively coordinates a diverse expert pool and a structurally-aware gating mechanism for principled, fine-grained routing.
    \item Through theoretical analysis, we address the training instability of naively stacking heterogeneous experts. We demonstrate that our backbone-residual design, in concert with an information-theoretic diversity regularizer, enables a stable and effective unsupervised adaptive framework.
    \item Comprehensive experiments across 16 benchmarks for unsupervised node classification and few-shot learning validate ADaMoRE's state-of-the-art performance and superior generalization. Our architecture demonstrates superior training stability, converging significantly faster and to a more stable, lower loss than naive MoE stacking For instance, it not only outperforms the next-best method by over 13\% in accuracy but also exhibits significantly higher training efficiency, effectively unlocking the potential of unsupervised MoE for robust graph representation learning.
\end{itemize}

%% file: 2_Preliminary.tex
\vspace{-7pt}
\section{Preliminary}

\subsection{Background}
\label{subsec:preliminary}
\paragraph{Graph} An undirected graph is $\mathcal{G} = (\mathcal{V}, \mathcal{E})$ with $N$ nodes $v_i \in \mathcal{V}$ and edges $(v_i, v_j) \in \mathcal{E}$. The adjacency matrix is $\mathbf{A} \in \{0, 1\}^{N \times N}$. We use $\hat{\mathbf{A}} = \mathbf{A} + \mathbf{I}_N$ for self-loops, with $\hat{\mathbf{D}}$ as its diagonal degree matrix ($\hat{\mathbf{D}}_{ii} = \sum_j \hat{A}_{ij}$). The symmetrically normalized adjacency matrix is $\tilde{\mathbf{A}} = \hat{\mathbf{D}}^{-1/2} \hat{\mathbf{A}} \hat{\mathbf{D}}^{-1/2}$. Node features are $\mathbf{x}_i \in \mathbb{R}^F$, a column vector of dimension $F$, forming the matrix $\mathbf{X} \in \mathbb{R}^{N \times F}$ where the $i$-th row is $\mathbf{x}_i^\top$. We use $\hat{\cdot}$ to denote matrices with self-loops and $\tilde{\cdot}$ for normalized matrices.

\paragraph{Graph Neural Networks}
A Graph Neural Network (GNN) learns node representations by iteratively aggregating information from its local neighborhood $\mathcal{N}(i)$. A general message-passing scheme for a node $v_i$ can be expressed as:
\begin{equation}
    \mathbf{h}^{(l+1)}_i = \text{UPDATE}\left(\mathbf{h}^{(l)}_i, \text{AGGREGATE}\left(\left\{\mathbf{h}^{(l)}_j \mid v_j \in \mathcal{N}(i)\right\}\right)\right).
\label{eq:message_passing}
\end{equation}
From a Graph Signal Processing perspective, this aggregation can be interpreted as applying a graph filter. For instance, a \textbf{Low-Pass Filter (LPF)}, as used in \textbf{SGC}~\cite{wu2019simplifyinggraphconvolutionalnetworks}, smooths features across neighbors and can be expressed as $\mathbf{H}_{LPF} = \tilde{\mathbf{A}}\mathbf{X}$. Correspondingly, a \textbf{High-Pass Filter (HPF)} accentuates feature differences and can be formulated as $\mathbf{H}_{HPF} = \tilde{\mathbf{L}}\mathbf{X}$, where $\tilde{\mathbf{L}} = \mathbf{I}_N - \alpha \tilde{\mathbf{A}}$ is based on the graph Laplacian~\cite{GREET}; we term this operator \textbf{LapSGC} in our context.

\paragraph{Structural Roles and Embeddings}
A node's structural role describes its topological function within the graph, independent of its features. To enable a GNN to recognize these roles, they can be quantified through structural embeddings. We adopt an approach based on random walks~\cite{dwivedi2022graphneuralnetworkslearnable}, where a node $v_k$'s structural embedding $\mathbf{s}_k$ is formed by its $p$-step return probabilities derived from the random walk transition matrix $\mathbf{T} = \hat{\mathbf{D}}^{-1}\hat{\mathbf{A}}$:
\begin{equation}
    \mathbf{s}_k = [T_{kk}, T^2_{kk}, \dots, T^{d_s}_{kk}]^\top.
\label{eq:structural_embedding}
\end{equation}
This construction represents a low-complexity usage of the random walk matrix, as it considers only the probability of a walk starting at a node returning to itself. This provides a feature-independent signature of a node's local topology, which is essential for guiding the adaptive mechanisms in our framework.

\subsection{Motivation}
\label{sec:motivation}
To handle the diverse nature of real-world graphs, a promising direction is to employ a MoE framework with a pool of heterogeneous GNNs, each possessing a unique inductive bias. However, our core finding is that naively stacking such experts leads to inherent training instability. This instability can be formally explained by the prohibitively high sample complexity of such a system.

\begin{lemma}[Catastrophic Sample Complexity of Naive Heterogeneous MoE]
\label{lemma:naive_moe_complexity}
\textit{Consider a naive MoE with a gating network $g \in \mathcal{G}$ and heterogeneous experts $\{E_k \in \mathcal{F}_k\}_{k=1}^K$. Its function space $\mathcal{F}_{Naive-hetero}$ is a gated union of disparate function classes. According to statistical learning theory, the complexity of this underlying union, $\cup_{k=1}^K \mathcal{F}_k$, has a Rademacher complexity upper-bounded by:}
\begin{equation}
    \mathcal{R}_n\left(\bigcup_{k=1}^K \mathcal{F}_k\right) \le \sqrt{\frac{2\log(K)}{n}} + \sum_{k=1}^K \mathcal{R}_n(\mathcal{F}_k).
\end{equation}
\textit{This already-high complexity, further amplified by the gating network's complexity $\mathcal{R}_n(\mathcal{G})$, leads to an enormous overall complexity $\mathcal{R}_n(\mathcal{F}_{Naive-hetero})$, resulting in a prohibitively high sample complexity:}
\begin{equation}
    N_{Naive}(\epsilon, \delta) \ge \Omega\left( \frac{\mathcal{R}_n(\mathcal{F}_{Naive-hetero})^2}{\epsilon^2} \right).
\end{equation}
\end{lemma}

\begin{remark}
\upshape
The high complexity stated in Lemma~\ref{lemma:naive_moe_complexity} arises from the conflicting inductive biases among heterogeneous experts. This is in stark contrast to a homogeneous MoE, where all experts belong to a single class $\mathcal{F}_{homo}$. A homogeneous MoE's function space is the convex hull $\text{conv}(\mathcal{F}_{homo})$. As a standard result from statistical learning theory, the complexity of a convex hull is equal to the original class, $\mathcal{R}_n(\text{conv}(\mathcal{F}_{homo})) = \mathcal{R}_n(\mathcal{F}_{homo})$. This theoretical difference explains why a naive heterogeneous MoE is inherently unstable under weak supervision.
\end{remark}

To overcome the fundamental challenge defined by Lemma~\ref{lemma:naive_moe_complexity}, we propose a novel \textbf{Backbone-Residual Architecture} designed to radically reduce the effective function space complexity.

\begin{lemma}[Sample Complexity Reduction via Backbone-Residual Architecture]
\label{lemma:sample_complexity_reduction}
\textit{A Backbone-Residual MoE decomposes the learning problem. By leveraging a powerful backbone to approximate the primary component of the target function, it allows the residual experts to learn a simplified target within a much smaller \textbf{effective function space} $\mathcal{F}_{res, eff}$. The sample complexity for this residual task, $N_{Res}(\epsilon, \delta)$, is consequently much lower:}
\begin{equation}
    N_{Res}(\epsilon, \delta) \le C \cdot \left( \frac{\mathcal{R}_n(\mathcal{F}_{res, eff})}{\mathcal{R}_n(\mathcal{F}_{Naive-hetero})} \right)^2 N_{Naive}(\epsilon, \delta) \ll N_{Naive}(\epsilon, \delta).
\end{equation}
\end{lemma}
Please refer to Appendix~\ref{ap:justification_lemma_2} for detailed proof. The efficacy of the Backbone-Residual Architecture hinges on a backbone powerful enough to satisfy its underlying assumptions of providing a high-quality approximation. To this end, we select and validate our backbone based on the following proposition.

\begin{proposition}[Efficacy of Guided Complementary Filtering as a Backbone]
\label{prop:backbone_efficacy}
\textit{The \textbf{Guided Complementary Filtering} paradigm, which utilizes a Low-Pass Filter and a High-Pass Filter to capture cohesive and dispersive signals, serves as a powerful and robust backbone. Our theoretical and empirical analyses (Please refer to Appendix~\ref{ap:support_prop_backbone}) confirm that its efficacy is critically contingent on the \textbf{accuracy} and \textbf{numerical distinctiveness} of its guiding weights.}
\end{proposition}

In summary, our work first formalizes the instability of naive heterogeneous MoE through the lens of sample complexity, then proposes a theoretically-grounded architectural solution, and finally instantiates it with a powerful backbone whose critical success factors we have identified and validated.

%% file: 3_Method.tex
\section{The ADaMoRE Framework}
\label{method}

\subsection{Overall Architecture}
\label{subsec:overview}
\begin{figure*}[t]
    \centering
    \captionsetup{skip=3pt}
    \includegraphics[width=1.0\textwidth]{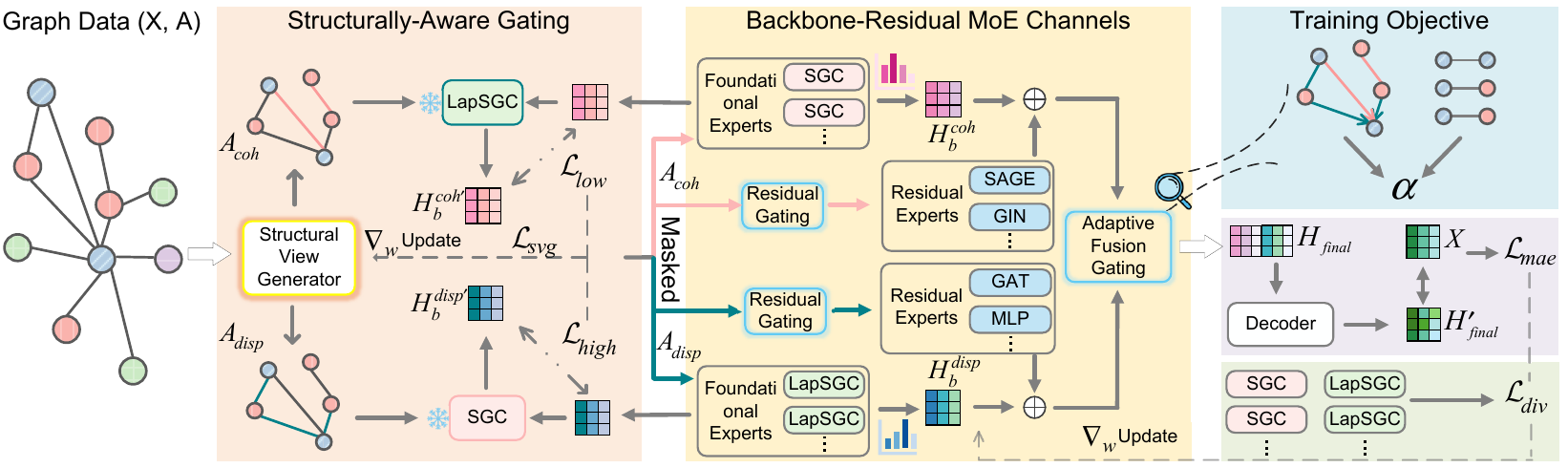}
    \caption{Overview of the proposed ADaMoRE framework. A Structurally-Aware Gating module first generates two complementary structural views. These views are then processed by parallel Backbone-Residual MoE Channels. A final Adaptive Fusion Gating module combines the enhanced outputs. The framework is trained via a step-wise optimization, alternating between a primary MAE objective for the main model and a dedicated cross-filter reconstruction loss for the view generator.}
    \label{fig:framework_overview}
    \Description{An architectural diagram of the ADaMoRE framework. Data flows from left to right. It starts with an input graph feeding into a 'Structural View Generator', which outputs two distinct graph views ('Cohesive' and 'Dispersive'). These two views are then processed by two parallel, identical 'Hybrid Expert Channels'. Each channel's internal structure shows a main 'MoE Backbone' path being added to the combined output of a 'Residual Expert Pool'. Finally, the outputs of the two channels are fed into an 'Adaptive Channel Fusion' module, which produces the final node embeddings. A large feedback arrow at the bottom indicates that the entire system is trained end-to-end.}
\end{figure*}

The ADaMoRE framework, illustrated in Figure~\ref{fig:framework_overview}, is an unsupervised framework that learns adaptive graph representations by dynamically composing a diverse set of GNN experts. Its core process begins with a Structurally-Aware Gating module, which analyzes the graph's local connectivity patterns to produce two complementary structural views. These views are then processed by parallel Backbone-Residual MoE Channels, each featuring our proposed backbone-residual architecture. Within each channel, a sparse MoE of foundational experts provides a stable backbone representation, which is subsequently enhanced by a dense ensemble of diverse, advanced GNNs via residual connections. Finally, an Adaptive Fusion Gating module intelligently combines the enhanced outputs of the two channels. It employs a node-specific coefficient, derived from the initial structural analysis, to perform a fine-grained, context-aware fusion, yielding the final node embeddings. Subsequent subsections provide detailed descriptions of each component and the overall training methodology.

\subsection{Structurally-Aware Gating}
\label{subsec:structurally_aware_gating}
The Structurally-Aware Gating module is the core component in ADaMoRE for perceiving local graph structures and generating adaptive signals. Architecturally, it consists of a learnable neural network to analyze connectivity patterns and a set of parameter-free filters to enable its training. The module performs two primary functions: (1) generating two complementary structural views of the graph, and (2) computing a node-specific coefficient for the final channel fusion.

The learnable component is an MLP, $MLP_\theta$, tasked with quantifying local connectivity patterns. For each edge $e = (v_i, v_j)$, it processes a joint representation $\mathbf{z}_{ij} \in \mathbb{R}^{2(F + d_s)}$ formed by concatenating the features of the incident nodes ($\mathbf{x}_i, \mathbf{x}_j$) and their pre-computed structural embeddings (Section~\ref{subsec:preliminary}). For undirected graphs, the edge logit $\ell_e$ averages the MLP outputs for both directions:
\begin{equation}
    \ell_e = \frac{1}{2} \left(MLP_\theta(\mathbf{z}_{ij}) + MLP_\theta(\mathbf{z}_{ji}) \right). \label{eq:edge_logit}
\end{equation}
To obtain differentiable weights suitable for end-to-end training, we employ the Gumbel-Sigmoid reparameterization on $\ell_e$. This technique provides a differentiable approximation for sampling binary-like edge weights, allowing gradients to flow back to the MLP. Furthermore, it encourages the model to learn highly discriminative weights by pushing the output towards 0 or 1, especially with a low temperature $\tau$. The resulting weight $w_e$ is computed as:
\begin{equation}
    w_e = \sigma \left( \frac{\ell_e + g_1 - g_2}{\tau} \right). \label{eq:gumbel_sigmoid}
\end{equation}
These learned edge weights, collected in a matrix $\mathbf{W}$, are used to generate two complementary views of the graph structure by modulating the original adjacency matrix $\mathbf{A}$:
\begin{equation}
    \mathbf{A}_{coh} = \mathbf{A} \odot \mathbf{W}, \quad \mathbf{A}_{disp} = \mathbf{A} \odot (\mathbf{1} - \mathbf{W}),
    \label{eq:graph_views}
\end{equation}
where $\odot$ is the Hadamard product. The \textit{Cohesive View} $\mathbf{A}_{coh}$ emphasizes community-like patterns, while the \textit{Dispersive View} $\mathbf{A}_{disp}$ highlights boundary-like patterns.

To facilitate the training of MLP without direct supervision, the module also incorporates two parameter-free graph filters. These filters are the core operators in a dedicated self-supervised training objective for this module, termed \textit{cross-filter reconstruction}. The detailed formulation of this training objective is provided in Section~\ref{subsec:training_objective}.

\subsection{The Sparse MoE Backbone}
\label{subsec:moe_backbone}
The backbone of our framework is designed as a sparse MoE system. It consists of two parallel expert banks, each dedicated to processing one of the structural views generated by the Structurally-Aware Gating module.

Each expert bank is composed of $N_{exp}$ homogeneous, yet distinct, foundational experts. These experts share the same architecture (SGC or Lap-SGC) but differ in their configuration, such as the number of propagation hops $K_{hops}$. This allows the backbone to capture structural information at multiple scales. Let the set of experts for the cohesive channel be $\{\mathcal{E}_{coh,k}\}_{k=1}^{N_{exp}}$ and for the dispersive channel be $\{\mathcal{E}_{disp,k}\}_{k=1}^{N_{exp}}$.

To achieve adaptive and efficient computation, a sparse gating mechanism operates within each bank. For each node $v_i$, a lightweight gating network computes a routing score for every expert in the bank. A Top-K operation then selects the $K$ experts with the highest scores. The final backbone representation for a given channel (denoted by $c \in \{coh, disp\}$) is the weighted sum of the outputs from these selected experts:

\begin{equation}
    \mathbf{h}_{b, i}^{(c)} = \sum_{k \in \mathcal{T}_K(i,c)} g_{i,k}^{(c)} \mathcal{E}_{c,k}(\mathbf{X}, \mathbf{A}_{c})_i,
    \label{eq:moe_backbone_output}
\end{equation}

where $\mathcal{T}_K(i,c)$ is the index set of the top-$K$ experts for node $v_i$ in channel $c$, and $g_{i,k}^{(c)}$ is the normalized gating weight. This process yields two distinct backbone representations, $\mathbf{H}_{b}^{coh}$ and $\mathbf{H}_{b}^{disp}$, which serve as the stable base for the subsequent residual augmentation.

To ensure that all foundational experts are effectively utilized during training and to prevent expert collapse, we incorporate a load balancing loss, $\mathcal{L}_{load}$. The detailed formulation of this loss is provided in Section~\ref{subsec:training_objective}.

\subsection{Residual Expert Augmentation}
\label{subsec:residual_augmentation}

While the Sparse MoE Backbone provides a stable and adaptive foundational representation, its homogeneous experts primarily capture dominant structural signals. To model the more complex, non-linear patterns and diverse computational capabilities required for challenging graphs, we introduce a supplementary mechanism termed \textbf{Residual Expert Augmentation}.

The augmentation is realized by a dense ensemble of \textbf{heterogeneous experts}, $\{\mathcal{E}_{r,k}\}_{k=1}^{N_r}$. This pool is intentionally composed of GNNs to provide a rich set of computational perspectives. Unlike the sparse backbone, all $N_r$ residual experts are activated for each node. The outputs of these experts are combined via a weighted sum with learnable scaling factors $\{\gamma_k\}_{k=1}^{N_r}$ to form a unified residual signal, $\mathbf{h}_{r}^{(c)}$. This signal is then added to the backbone's output $\mathbf{h}_{b}^{(c)}$ for each channel $c \in \{coh, disp\}$ to yield the final enhanced representation:
\begin{equation}
    \mathbf{h}_{enhanced}^{(c)} = \mathbf{h}_{b}^{(c)} + \mathbf{h}_{r}^{(c)} = \mathbf{h}_{b}^{(c)} + \sum_{k=1}^{N_r} \gamma_k \cdot \mathcal{E}_{r,k}(\mathbf{X}, \mathbf{A}_{c}).
    \label{eq:residual_combination}
\end{equation}

To ensure these powerful, heterogeneous experts learn complementary, non-redundant functions, their training is guided by an information-theoretic diversity regularizer, denoted as $\mathcal{L}_{div}$. This objective explicitly penalizes the similarity between the outputs of different experts, thereby enforcing functional specialization. The detailed formulation of $\mathcal{L}_{div}$ is provided in Section~\ref{subsec:training_objective}.

\subsection{Adaptive Fusion Gating}
\label{subsec:adaptive_fusion}
The final component of the ADaMoRE framework is the Adaptive Fusion Gating, which intelligently blends the representations from the two parallel Backbone-Residual MoE Channels: the enhanced cohesive representation $\mathbf{H}_{e}^{(coh)}$ and the enhanced dispersive representation $\mathbf{H}_{d}^{(disp)}$. To achieve this, the module computes a node-specific coefficient $\alpha_i \in [0, 1]$ for each node $v_i$ to determine the optimal balance between the two signals.

The computation of $\alpha_i$ is guided by both structural and semantic cues derived from the graph context. It begins with an initial estimate, $\alpha_{i,init}$, which is the average of a structural score and a semantic score. The structural score, $\alpha_{i,struct}$, is calculated based on the statistics of the learned edge weights $\mathbf{W}$ incident to node $v_i$ from the Structurally-Aware Gating module. The semantic score, $\alpha_{i,sem}$, is derived from the local embedding consistency within the cohesive channel, as this channel is expected to best capture community-centric smoothness:
\begin{equation}
    \alpha_{i,sem} = \frac{1}{|\mathcal{N}(i)|} \sum_{j \in \mathcal{N}(i)} \cos\left(\mathbf{h}_{e,i}^{(coh)}, \mathbf{h}_{e,j}^{(coh)}\right).
    \label{eq:semantic_score}
\end{equation}
The initial estimate is then refined via a single step of message passing to smooth the coefficients across the local neighborhood:
\begin{equation}
    \alpha_i = \sum_{j \in \mathcal{N}(i) \cup \{i\}} \tilde{\mathbf{A}}_{i,j} \left( \frac{\alpha_{j,struct} + \alpha_{j,sem}}{2} \right).
    \label{eq:gating_propagation}
\end{equation}
The resulting coefficients, collected as a vector $\bm{\alpha} \in \mathbb{R}^N$ and clamped to $[0, 1]$, guide the final representation generation. This is achieved by first scaling each channel's output and then concatenating the results to form the final, comprehensive node representation $\mathbf{H}_{final} \in \mathbb{R}^{N \times 2d_e}$:
\begin{equation}
    \mathbf{H}_{final} = \operatorname{concat} \left( \bm{\alpha} \odot \mathbf{H}_{e}^{(coh)}, (\mathbf{1} - \bm{\alpha}) \odot \mathbf{H}_{e}^{(disp)} \right),
    \label{eq:final_representation}
\end{equation}
where $\odot$ denotes element-wise multiplication and $\mathbf{1}$ is the all-ones vector.

\subsection{Training Objective}
\label{subsec:training_objective}
The ADaMoRE framework is trained end-to-end via a principled, multi-objective unsupervised paradigm. To ensure stable and effective learning for the different components, we employ a step-wise optimization strategy.

\noindent \textbf{Primary Objective: Masked Representation Reconstruction.}
The primary learning signal for the entire framework is a Masked Autoencoder (MAE) task, which trains the model to produce rich, informative representations. During each training step, we randomly select a subset of nodes $\mathcal{M}$ and mask their input features. The full graph with the unmasked features is processed by the ADaMoRE framework to produce the final node embeddings $\mathbf{H}_{final}$. A shared MLP decoder then attempts to reconstruct the original features $\mathbf{x}_i$ for only the masked nodes from their final embeddings $\mathbf{h}_{final, i}$. The reconstruction loss, $\mathcal{L}_{mae}$, is defined as the scaled cosine error over the masked set:
\begin{equation}
    \mathcal{L}_{mae} = \frac{1}{|\mathcal{M}|} \sum_{i \in \mathcal{M}} \left( 1 - \cos(\text{Decoder}(\mathbf{h}_{final, i}), \mathbf{x}_i)^{\gamma} \right),
    \label{eq:reconstruction_loss}
\end{equation}
where $\gamma > 0$ is a scaling exponent. This objective drives the model to learn comprehensive embeddings that retain sufficient information to infer missing node attributes from the graph context.

\noindent \textbf{Regularization for Experts.}
To enable the stable training of our complex MoE architecture, two key regularizers are employed.
First, to prevent expert collapse within the sparse MoE backbone, we use a standard Load Balancing Loss $\mathcal{L}_{load}$~\cite{adaptivity}. This loss encourages the gating network to distribute nodes evenly across the foundational experts.
Second, to enforce functional specialization among the experts, we integrate an Information-Theoretic Diversity Regularizer $\mathcal{L}_{div}$. This objective penalizes the similarity between the outputs of different residual experts, thereby compelling them to learn complementary, non-redundant functions. We implement this using Centered Kernel Alignment (CKA)~\cite{kornblith2019similarityneuralnetworkrepresentations}, a robust measure of representation similarity. For the outputs of two experts, $\mathbf{E}_i$ and $\mathbf{E}_j$, the CKA-based diversity loss is formulated as:
\begin{equation}
    \mathcal{L}_{div}(\mathbf{E}_i, \mathbf{E}_j) = \text{CKA}(\mathbf{E}_i, \mathbf{E}_j) = \frac{\text{HSIC}(\mathbf{E}_i, \mathbf{E}_j)}{\sqrt{\text{HSIC}(\mathbf{E}_i, \mathbf{E}_i) \cdot \text{HSIC}(\mathbf{E}_j, \mathbf{E}_j)}},
    \label{eq:cka_loss}
\end{equation}
where $\text{HSIC}(\mathbf{E}_i, \mathbf{E}_j) = \text{tr}(\mathbf{K}_i^c \mathbf{K}_j^c)$ is the Hilbert-Schmidt Independence Criterion, and $\mathbf{K}^c$ denotes a centered kernel matrix computed from the expert's output representations.

\noindent \textbf{Self-Supervision for Structural View Generation.}
The learnable MLP within the Structurally-Aware Gating module is trained independently via a dedicated self-supervised objective, $\mathcal{L}_{svg}$, termed \textbf{Cross-Filter Reconstruction}. This strategy creates an adversarial-like task that guides the MLP to produce meaningful edge weights without any labels. Specifically, it comprises two objectives. First, a  parameter-free LPF attempts to reconstruct the output of the high-pass backbone expert, $\mathbf{H}_{b}^{(disp)}$, using the dispersive view $\mathbf{A}_{disp}$:
\begin{equation}
    \mathcal{L}_{l \to h} = \frac{1}{N} \sum_{i=1}^{N} \left( 1 - \text{cos}\left(LPF(\mathbf{H}_{b}^{(disp)}, \mathbf{A}_{disp})_i, \mathbf{h}_{b,i}^{(disp)}\right)^{\gamma_{svg}} \right).
    \label{eq:loss_l_to_h}
\end{equation}
Symmetrically, a fixed HPF attempts to reconstruct the output of the low-pass backbone expert, $\mathbf{H}_{b}^{(coh)}$, using the cohesive view $\mathbf{A}_{coh}$, yielding a loss $\mathcal{L}_{h \to l}$. The total loss for the view generator is $\mathcal{L}_{svg} = \mathcal{L}_{l \to h} + \mathcal{L}_{h \to l}$. By minimizing this combined loss, the MLP learns to assign edge weights that make these "mismatched" reconstruction tasks maximally difficult, which in turn means the weights have successfully captured the underlying structural patterns.

\noindent \textbf{Overall Training Process and Extension}
The main ADaMoRE model is trained by minimizing a composite objective that combines the primary reconstruction loss with the two regularization terms:
\begin{equation}
    \mathcal{L}_{\text{ADaMoRE}} = \mathcal{L}_{mae} + \lambda_{load}\mathcal{L}_{load} + \lambda_{div}\mathcal{L}_{div},
    \label{eq:total_loss_adamore}
\end{equation}
where $\lambda_{load}$ and $\lambda_{div}$ are balancing coefficients. The overall framework is trained by alternating between two steps: (1) updating the parameters of the Structurally-Aware Gating's MLP using $\mathcal{L}_{svg}$, and (2) updating the parameters of the main ADaMoRE model using $\mathcal{L}_{\text{ADaMoRE}}$.

Furthermore, our pre-trained framework can be seamlessly adapted for \textbf{Few-Shot Learning}. After unsupervised pre-training, a classification head is added, and the entire model is fine-tuned on the few labeled examples by minimizing a combined loss that includes a standard classification objective: $\mathcal{L}_{few-shot} = \mathcal{L}_{\text{ADaMoRE}} + \lambda_{cls}\mathcal{L}_{cls}$.

\begin{table*}[t]
\captionsetup{skip=3pt}
\caption{Performance comparison on the unsupervised node classification task (mean accuracy \% $\pm$ std. dev. over 10 runs). Best and second-best results are in \textbf{bold} and \underline{underlined}, respectively. The symbol "OOM" denotes out of memory.}
\label{tab:main_unsupervised_results_final}
\Description{A merged and final table showing unsupervised node classification accuracy on ten key datasets, categorized into five homophilic (Cora, Pubmed, Computers, CS, Ogbn-Arxiv) and five heterophilic (Cornell, Texas, Wisconsin, Roman-Empire, Minesweeper). The proposed ADaMoRE model demonstrates highly competitive performance across this diverse set.}
\centering
\setlength{\tabcolsep}{3pt} 
\begin{tabular}{@{}l|ccccc|ccccc@{}}
\toprule
& \multicolumn{5}{c|}{\textbf{Homophilic Datasets}} & \multicolumn{5}{c}{\textbf{Heterophilic Datasets}} \\
\cmidrule(r){2-6} \cmidrule(l){7-11}
Methods & Cora & Pubmed & Computers & CS & \makecell{Ogbn-\\Arxiv} & Cornell & Texas & Wisconsin & \makecell{Roman-\\Empire} & \makecell{Mine-\\sweeper} \\ \midrule
DGI\cite{DGI}     & $82.84\pm_{1.31}$ & $80.22\pm_{0.52}$ & $85.41\pm_{0.61}$ & $91.70\pm_{0.35}$ & $69.50\pm_{0.32}$ & $58.68\pm_{5.04}$ & $75.81\pm_{5.03}$ & $71.46\pm_{5.52}$ & $42.04\pm_{0.94}$ & $79.95\pm_{0.75}$ \\
MaskGAE\cite{MaskGAE} & $86.50\pm_{1.22}$ & $83.07\pm_{0.64}$ & $72.44\pm_{0.54}$ & $88.84\pm_{0.34}$ & OOM & $65.50\pm_{7.44}$ & $76.04\pm_{6.42}$ & $71.54\pm_{6.24}$ & $62.21\pm_{0.84}$ & $79.71\pm_{0.72}$ \\
GDAE\cite{GDAE}   & $87.37\pm_{1.14}$ & $83.70\pm_{0.69}$ & $72.74\pm_{0.55}$ & $89.35\pm_{0.28}$ & OOM & $66.32\pm_{8.17}$ & $76.84\pm_{6.68}$ & $72.16\pm_{6.55}$ & $62.90\pm_{0.67}$ & $79.66\pm_{0.68}$ \\
MVGRL\cite{MVGRL} & $86.78\pm_{1.02}$ & $86.76\pm_{0.55}$ & $87.19\pm_{0.45}$ & $91.07\pm_{0.31}$ & $70.10\pm_{0.33}$ & $61.26\pm_{4.01}$ & $73.01\pm_{5.52}$ & $72.82\pm_{4.51}$ & $46.68\pm_{0.73}$ & $80.83\pm_{0.85}$ \\
GRACE\cite{GRACE} & $85.26\pm_{1.41}$ & $82.13\pm_{0.65}$ & $80.67\pm_{0.65}$ & $91.46\pm_{0.42}$ & $70.11\pm_{0.25}$ & $45.85\pm_{5.51}$ & $72.15\pm_{4.53}$ & $73.00\pm_{5.21}$ & $50.24\pm_{1.02}$ & $79.99\pm_{0.81}$ \\
GBT\cite{GBT}     & $83.67\pm_{1.52}$ & $79.77\pm_{0.61}$ & $82.44\pm_{0.55}$ & $92.62\pm_{0.42}$ & $69.60\pm_{0.28}$ & $49.32\pm_{4.52}$ & $69.25\pm_{4.81}$ & $63.65\pm_{4.03}$ & $37.76\pm_{0.85}$ & $78.34\pm_{0.71}$ \\
BGRL\cite{BGRL}   & $82.32\pm_{1.63}$ & $85.36\pm_{0.51}$ & $87.63\pm_{0.41}$ & $91.85\pm_{0.34}$ & $\underline{70.45\pm_{0.35}}$ & $54.98\pm_{6.02}$ & $68.41\pm_{5.01}$ & $61.20\pm_{6.52}$ & $50.79\pm_{1.13}$ & $80.07\pm_{0.65}$ \\
DSSL\cite{DSSL}   & $85.35\pm_{1.12}$ & $86.09\pm_{0.63}$ & $82.44\pm_{0.51}$ & $92.16\pm_{0.35}$ & $69.45\pm_{0.28}$ & $69.28\pm_{6.01}$ & $72.73\pm_{4.02}$ & $79.26\pm_{3.51}$ & $59.69\pm_{0.75}$ & $79.79\pm_{0.73}$ \\
GREET\cite{GREET} & $84.49\pm_{1.76}$ & $86.89\pm_{0.45}$ & $86.43\pm_{0.46}$ & \underline{$93.35\pm_{0.42}$} & OOM & $70.00\pm_{5.34}$ & $80.79\pm_{4.41}$ & $82.16\pm_{4.51}$ & $63.89\pm_{1.34}$ & \underline{$81.76\pm_{0.88}$} \\
SP-GCL\cite{SP-GCL}& $86.56\pm_{1.51}$ & $85.45\pm_{0.45}$ & \underline{$88.72\pm_{0.65}$} & $91.03\pm_{0.41}$ & $68.40\pm_{0.32}$ & $70.69\pm_{6.51}$ & $76.83\pm_{5.82}$ & $77.10\pm_{5.53}$ & $50.79\pm_{0.95}$ & $79.75\pm_{0.64}$ \\
S3GCL\cite{S3GCL} & \underline{$87.57\pm_{0.96}$} & \underline{$87.68\pm_{0.44}$} & $87.73\pm_{0.59}$ & $92.67\pm_{0.41}$ & OOM & \underline{$73.33\pm_{7.34}$} & \underline{$82.50\pm_{5.86}$} & \underline{$82.40\pm_{6.85}$} & \underline{$64.55\pm_{0.75}$} & $80.37\pm_{0.81}$ \\
\rowcolor{mygray}
ADaMoRE& $\bm{88.93\pm_{1.53}}$ & $\bm{89.29\pm_{0.49}}$ & $\bm{89.66\pm_{0.67}}$ & $\bm{94.70\pm_{0.43}}$ & $\bm{71.23\pm_{0.29}}$ & $\bm{77.37
\pm_{4.21}}$ & $\bm{86.05\pm_{5.76}}$ & $\bm{85.30\pm_{3.41}}$ & $\bm{77.86\pm_{0.89}}$ & $\bm{85.02\pm_{0.68}}$ \\ \bottomrule
\end{tabular}
\end{table*}

%% file: 4_Experiments.tex
\section{Experiments}
\label{experiments}
This section empirically evaluates our proposed ADaMoRE framework. We assess its core representation quality and generalization capabilities through unsupervised node classification (Section~\ref{sec:exp_unsupervised}) and few-shot learning (Section~\ref{sec:exp_fewshot}). Additionally, we conduct a comprehensive analysis of the framework's core components, training stability, and computational efficiency in Section~\ref{ablation_and_time}. Further supplementary results, including performance on a downstream node clustering task (Appendix~\ref{ap:node_clustering}) and a hyperparameter sensitivity analysis (Appendix~\ref{ap:hyperparameter_sensitivity}), are provided in the appendix. All implementation details and model configurations are detailed in Appendix~\ref{ap:model_configurations}.

\subsection{Evaluation on Unsupervised Tasks}
\label{sec:exp_unsupervised}
This subsection evaluates ADaMoRE's effectiveness in learning comprehensive graph representations for unsupervised node classification.

\noindent \textbf{Datasets.}
We evaluate our method on 10 node classification benchmarks covering both homophilic and heterophilic graphs. Specifically, for graphs with homophily, we adopt three widely-used citation networks: Cora and Pubmed~\cite{cora}, and the large-scale Ogbn-Arxiv~\cite{mikolov2013distributedrepresentationswordsphrases}; one co-purchase network: Computers~\cite{shchur2019pitfallsgraphneuralnetwork}; and one co-authorship network: CS~\cite{shchur2019pitfallsgraphneuralnetwork}. For graphs with heterophily, we adopt three webpage datasets: Cornell, Texas, and Wisconsin~\cite{single_graphs}; one Wikipedia-derived network: Roman-Empire~\cite{platonov2024criticallookevaluationgnns}; and one synthetic dataset: Minesweeper~\cite{platonov2024criticallookevaluationgnns}. The detailed statistics of these datasets are summarized in Table~\ref{tab:un_dataset_statistics} in Appendix~\ref{ap:dataset_details}.

\noindent \textbf{Baselines.}
We compare ADaMoRE against state-of-the-art unsupervised graph representation learning baselines. These encompass \textbf{Graph Auto-Encoder Methods} like MaskGAE\cite{MaskGAE} and GDAE\cite{GDAE}; popular \textbf{Graph Contrastive Learning Methods} such as DGI\cite{DGI}, GMI\cite{GMI}, MVGRL\cite{MVGRL}, GRACE\cite{GRACE}, GBT\cite{GBT}, BGRL\cite{BGRL}, DSSL\cite{DSSL} and SP-GCL\cite{SP-GCL}; and recent \textbf{Adaptive GNN Methods}, including GREET\cite{GREET} and S3GCL\cite{S3GCL}.

\noindent \textbf{Training and Evaluation.}
Following standard protocols, node embeddings are first learned unsupervisedly. Subsequently, a linear classifier is trained on these frozen embeddings to evaluate representation quality. We report the average accuracy and standard deviation over ten runs.

\noindent \textbf{Main Results.}
The results in Table~\ref{tab:main_unsupervised_results_final} demonstrate the robust and superior performance of ADaMoRE, which consistently achieves state-of-the-art results across a diverse range of graph benchmarks, showcasing its strong generalization ability. Unlike baselines that are often confined to a single GNN encoder or a fixed set of filters, ADaMoRE's core advantage stems from its principled approach to managing a diverse pool of heterogeneous experts in a fully unsupervised setting. Our backbone-residual design, stabilized by an information-theoretic diversity regularizer, successfully addresses the training instability inherent in naive heterogeneous MoE. This enables ADaMoRE to effectively leverage the distinct computational capabilities of various GNNs, leading to a more powerful and nuanced adaptation to the complex structural patterns.

\subsection{Evaluation on Few-Shot Learning Tasks}
\label{sec:exp_fewshot}
This subsection evaluates the performance of ADaMoRE on few-shot learning benchmarks. These challenging tasks serve as a rigorous test of our method's generalization capabilities when adapting to new scenarios with minimal labeled data.

\noindent \textbf{Datasets.}
We evaluate ADaMoRE on six few-shot learning benchmarks covering both node and graph classification tasks. Specifically, for few-shot node classification, we adopt one webpage network: Wisconsin~\cite{single_graphs}; one Wikipedia-derived network: Squirrel~\cite{rozemberczki2021multiscaleattributednodeembedding}; and one citation network: Citeseer~\cite{cora}. For few-shot graph classification, we adopt three standard benchmarks: one protein structure dataset, PROTEINS~\cite{borgwardt2005protein}; one enzyme dataset, ENZYMES~\cite{wang2022faith}; and one chemical compound dataset, BZR~\cite{rossi2015network}. The detailed statistics of these datasets are summarized in Table~\ref{tab:few_shot_dataset_statistics} in Appendix~\ref{ap:dataset_details}.

\begin{table*}[t]
\captionsetup{skip=3pt}
\caption{Performance comparison on 1-shot node classification and graph classification tasks (mean accuracy \% $\pm$ std. dev. over 100 tasks). Best and second-best results are in \textbf{bold} and \underline{underlined}, respectively.}
\label{tab:main_fewshot_results_with_gmoe}
\Description{A comprehensive merged table showing few-shot learning performance. It is split into two sections for node and graph classification. The table compares ADaMoRE against a wide range of baselines, including specific GMoE variants, on six key datasets. The proposed ADaMoRE model demonstrates superior or highly competitive performance across all tasks and datasets shown.}
\centering
\setlength{\tabcolsep}{5pt} 
\begin{tabular}{@{}l|ccc|ccc@{}}
\toprule
& \multicolumn{3}{c|}{\textbf{Few-shot Node Classification}} & \multicolumn{3}{c}{\textbf{Few-shot Graph Classification}} \\
\cmidrule(r){2-4} \cmidrule(l){5-7}
Methods & Wisconsin & Squirrel & Citeseer & PROTEINS & ENZYMES & BZR \\ \midrule
GCN\cite{gcn}           & $21.39\pm_{6.56}$  & $20.00\pm_{0.29}$  & $31.27\pm_{4.53}$  & $51.66\pm_{10.87}$ & $19.30\pm_{6.36}$ & $45.06\pm_{16.30}$ \\
GAT\cite{GAT}           & $28.01\pm_{5.40}$  & $21.55\pm_{2.30}$  & $30.76\pm_{5.40}$  & $51.33\pm_{11.02}$ & $20.24\pm_{6.39}$ & $46.28\pm_{15.26}$ \\
H2GCN\cite{H2GCN}         & $23.60\pm_{4.64}$  & $21.90\pm_{2.15}$  & $26.98\pm_{6.25}$  & $53.81\pm_{8.85}$  & $19.40\pm_{5.57}$ & $50.28\pm_{12.13}$ \\
FAGCN\cite{FAGCN}         & $35.03\pm_{17.92}$ & $20.91\pm_{1.79}$  & $26.46\pm_{6.34}$  & $55.45\pm_{11.57}$ & $19.95\pm_{5.94}$ & $50.93\pm_{12.41}$ \\
DGI\cite{DGI}           & $28.04\pm_{6.47}$  & $20.00\pm_{1.86}$  & $45.00\pm_{9.19}$  & $50.32\pm_{13.47}$ & $21.57\pm_{5.37}$ & $49.97\pm_{12.63}$ \\
GraphCL\cite{GraphCL}       & $29.85\pm_{8.46}$  & $21.42\pm_{2.22}$  & $43.12\pm_{9.61}$  & $54.81\pm_{11.44}$ & $19.93\pm_{5.65}$ & $50.50\pm_{18.62}$ \\
DSSL\cite{DSSL}          & $28.46\pm_{10.31}$ & $20.94\pm_{1.88}$  & $39.86\pm_{8.60}$  & $52.73\pm_{10.98}$ & $23.14\pm_{6.71}$ & $49.04\pm_{8.75}$ \\
GraphACL\cite{GraphACL}      & $34.57\pm_{10.46}$ & $24.44\pm_{3.94}$  & $35.91\pm_{7.87}$  & $56.11\pm_{13.95}$ & $20.28\pm_{5.60}$ & $49.24\pm_{17.87}$ \\
GraphPrompt\cite{GraphPrompt}   & $31.48\pm_{5.18}$  & $21.22\pm_{1.80}$  & $45.34\pm_{10.53}$ & $53.61\pm_{8.90}$  & $21.85\pm_{6.17}$ & $50.46\pm_{11.46}$ \\
GraphPrompt+\cite{GraphPrompt+} & $31.54\pm_{4.54}$  & $21.24\pm_{1.82}$  & $45.23\pm_{10.01}$ & $54.55\pm_{12.61}$ & $21.85\pm_{5.15}$ & \underline{$53.26\pm_{14.99}$} \\
ProNoG\cite{ProNoG}        & \underline{$44.72\pm_{11.93}$} & \underline{$24.59\pm_{3.41}$} & \underline{$49.02\pm_{10.66}$} & \underline{$56.11\pm_{10.19}$} & $22.55\pm_{6.70}$ & $51.62\pm_{14.27}$ \\ \midrule
GMOE-GCN      & $29.32\pm_{8.42}$  & $20.52\pm_{0.85}$  & $48.90\pm_{9.30}$  & $55.59\pm_{10.36}$ & $22.33\pm_{3.56}$ & $52.64\pm_{13.68}$ \\
GMOE-SAGE     & $26.78\pm_{7.69}$  & $20.54\pm_{1.60}$  & $17.28\pm_{1.88}$  & $54.26\pm_{11.05}$ & $21.30\pm_{3.29}$ & $47.69\pm_{18.43}$ \\
GMOE-GIN      & $26.17\pm_{6.56}$  & $21.15\pm_{1.59}$  & $40.87\pm_{9.66}$  & $56.05\pm_{9.80}$ & \underline{$23.15\pm_{3.63}$} & $53.07\pm_{14.45}$ \\
\rowcolor{mygray}
ADaMoRE & $\bm{47.37\pm_{9.84}}$ & $\bm{26.83\pm_{5.24}}$ & $\bm{50.81\pm_{10.36}}$ & $\bm{57.33\pm_{11.43}}$ & $\bm{25.23\pm_{4.70}}$ & $\bm{55.92\pm_{13.98}}$ \\ \bottomrule
\end{tabular}
\end{table*}

\noindent \textbf{Baselines.}
We benchmark ADaMoRE against a comprehensive suite of state-of-the-art methods for few-shot learning. The baselines are categorized as follows: (1) \textbf{end-to-end Supervised GNNs}, such as GCN~\cite{gcn} and H2GCN~\cite{H2GCN}, trained directly on the few-shot labels; (2) \textbf{Graph Pre-training Models}, including GraphCL~\cite{GraphCL} and DSSL~\cite{DSSL}; (3) \textbf{Graph Prompt Learning Models}, such as GraphPrompt~\cite{GraphPrompt} and ProNoG~\cite{ProNoG}; and critically, (4) several \textbf{GMoE} variants~\cite{GMoE} to specifically evaluate our architectural contributions.

\noindent \textbf{Training and Evaluation Protocol.}
We evaluate on both node and graph classification under a $k$-shot setting, where only $k$ labeled instances per class are used for adaptation. For evaluation, we follow the standard prototype-based protocol~\cite{graph_few_shot_AM,liu2021relative}: class prototypes are computed from the $k$ labeled instances, and a test instance is assigned the label of its nearest prototype in the embedding space. In the main paper, we focus on the most challenging 1-shot setting ($k=1$), while the results for settings with more shots are provided in Appendix~\ref{ap:extended_few_shot}.

\noindent \textbf{Main Results.}
Table~\ref{tab:main_fewshot_results_with_gmoe} confirms the strong generalization ability of ADaMoRE in few-shot settings. Such robust results across both node and graph-level tasks underscore the effectiveness of the versatile and adaptive representations learned by ADaMoRE. This positions our framework as a potent and reliable solution for diverse graph-based tasks, particularly when operating under the constraints of limited supervision. Unlike other pre-training paradigms that may learn more generic embeddings, ADaMoRE's design, which combines a stable backbone with a functionally specialized residual expert pool, produces richer representations that are particularly effective for rapid adaptation from minimal data. The superior performance against various GMoE variants further validates our architectural choices, demonstrating that the backbone-residual design is a more effective and principled approach for constructing a powerful unsupervised Graph MoE.

\subsection{Ablation Studies and Analysis}
\label{ablation_and_time}
\noindent\textbf{Ablation Studies.}
Ablation studies, presented in Table~\ref{tab:ablation}, affirm the essential contribution of each component in the ADaMoRE framework. The results first underscore the efficacy of our core architectural innovations: removing the \textbf{Residual Experts} or training without the \textbf{Diversity Regularizer} both lead to a clear performance drop, validating the necessity of the backbone-residual design and the principled specialization it enables. The adaptive mechanisms are also shown to be critical. Disabling the \textbf{View Generator}, which forces both channels to process the same raw graph structure, or relying on a single-view pathway (\textbf{Cohesive/Dispersive View Only}) results in a significant performance decline, highlighting the importance of providing distinct structural perspectives. Finally, replacing the \textbf{Adaptive Fusion} with a static \textbf{NaiveConcat} baseline is suboptimal, confirming the benefit of a fine-grained, node-specific fusion strategy. Furthermore, we demonstrate the framework's robustness to imperfect guiding signals in Appendix~\ref{ap:weight_noise}, where performance degrades gracefully even when significant noise is introduced.

\begin{table}[h]
\captionsetup{skip=3pt}
\caption{Ablation study on the core components of ADaMoRE (mean accuracy \% $\pm$ std. dev.).}
\label{tab:ablation}
\centering
\resizebox{\columnwidth}{!}{%
\begin{tabular}{lcccc}
\toprule
Model Variant & Cora & Pubmed & Cornell & \makecell{Roman-Empire} \\
\midrule
\rowcolor{mygray}
\textbf{ADaMoRE (Full Model)} & $\bm{88.48\pm_{1.45}}$ & $\bm{89.29\pm_{0.49}}$ & $\bm{77.36\pm_{6.52}}$ & $\bm{77.86\pm_{0.85}}$ \\
\midrule
w/o Residual Experts & $87.12\pm_{1.58}$ & $88.05\pm_{0.53}$ & $74.15\pm_{6.81}$ & $75.21\pm_{0.98}$ \\
w/o Diversity Regularizer & $87.55\pm_{1.62}$ & $88.31\pm_{0.61}$ & $75.03\pm_{6.75}$ & $76.10\pm_{1.12}$ \\
w/o View Generator & $84.31\pm_{1.75}$ & $86.50\pm_{0.73}$ & $71.05\pm_{7.02}$ & $68.56\pm_{1.34}$ \\
w/o Adaptive Fusion & $86.95\pm_{1.65}$ & $88.13\pm_{0.58}$ & $75.81\pm_{6.60}$ & $74.98\pm_{1.05}$ \\
\midrule
NaiveConcat & $85.50\pm_{1.82}$ & $87.36\pm_{0.81}$ & $69.53\pm_{5.03}$ & $73.79\pm_{1.24}$ \\
Cohesive View Only & $86.13\pm_{1.61}$ & $87.46\pm_{0.62}$ & $58.86\pm_{7.04}$ & $52.35\pm_{1.53}$ \\
Dispersive View Only & $74.34\pm_{2.02}$ & $83.83\pm_{1.03}$ & $72.16\pm_{6.81}$ & $71.83\pm_{1.02}$ \\
\bottomrule
\end{tabular}
}
\end{table}

\begin{figure*}[t]
    \centering
    \captionsetup{skip=3pt}
    \begin{minipage}[t]{0.32\textwidth}
        \centering
        \includegraphics[width=\linewidth]{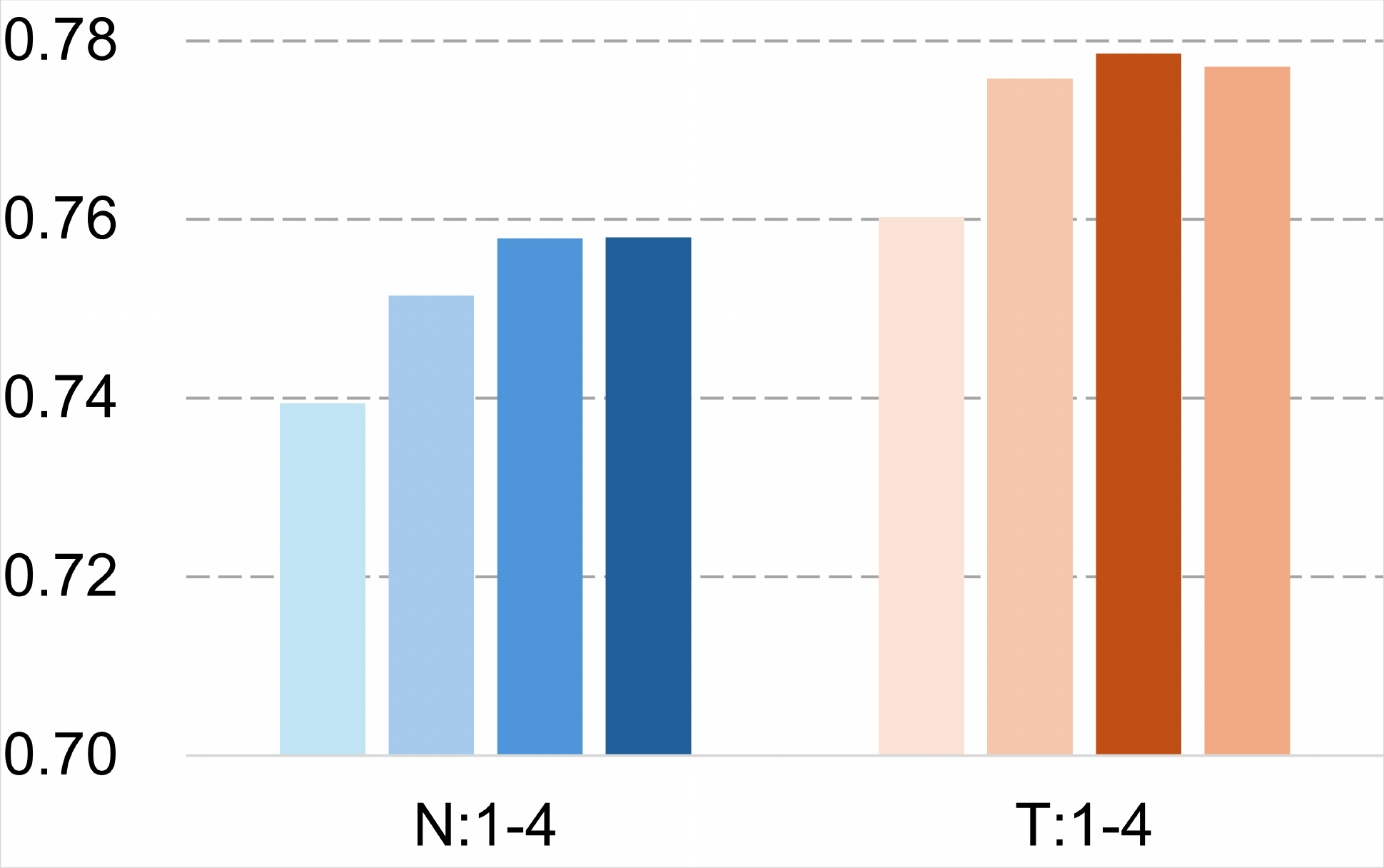}
        \subcaption{Ablation on expert number and diversity.}
        \label{fig:ablation_experts}
    \end{minipage}
    \hfill
    \begin{minipage}[t]{0.32\textwidth}
        \centering
        \includegraphics[width=\linewidth]{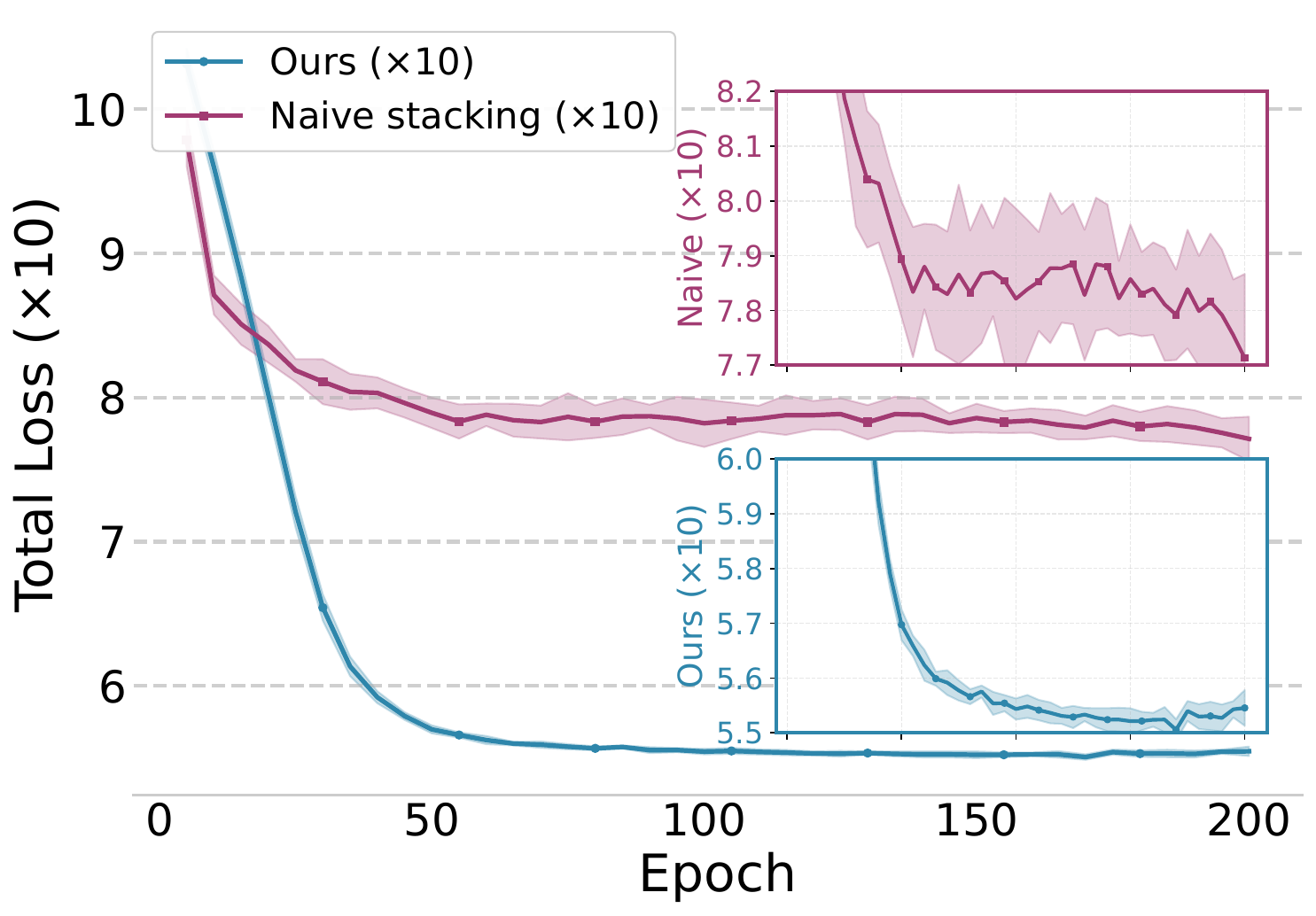}
        \subcaption{Training stability vs. naive stacking.}
        \label{fig:training_curve}
    \end{minipage}
    \hfill
    \begin{minipage}[t]{0.32\textwidth}
        \centering
        \includegraphics[width=\linewidth]{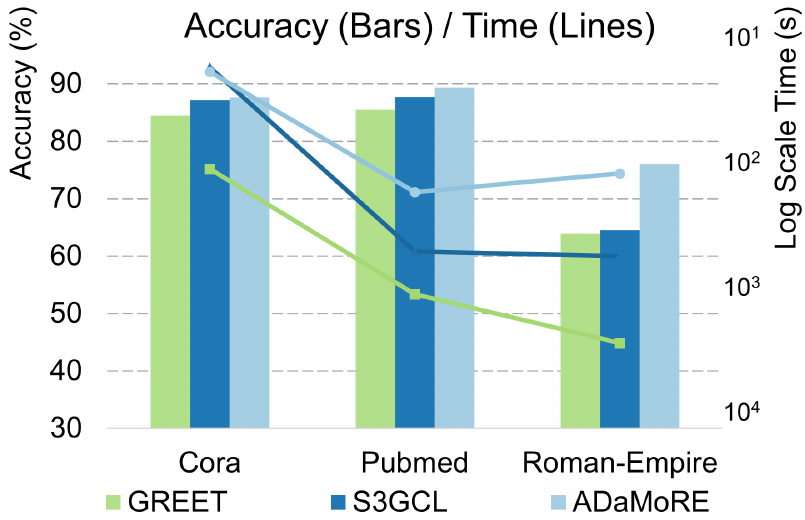}
        \subcaption{Performance vs. efficiency.}
        \label{fig:efficiency_comparison}
    \end{minipage}
    
    \caption{Comprehensive analysis of ADaMoRE, demonstrating (a) the benefits of expert diversity, (b) the superior training stability of our backbone-residual architecture, and (c) its compelling balance of high accuracy and efficiency.}
    \label{fig:comprehensive_analysis}
    \Description{A composite figure with three plots. Subfigure (a) shows ablation results for expert design. Subfigure (b) compares the training stability of our method against a naive baseline. Subfigure (c) shows the performance and efficiency trade-off against other methods.}
\end{figure*}

\noindent \textbf{Effect of MoE Design and Training Stability.}
Our architectural principles are validated by a series of targeted analyses, presented in Figure~\ref{fig:comprehensive_analysis}. First, we examine the MoE design (Figure~\ref{fig:comprehensive_analysis}(a)). The results show that performance consistently improves both when increasing the number of foundational experts in the backbone (N) and when progressively adding diverse residual experts (T). This empirically validates that a larger, more diverse expert pool, managed by our framework's mechanisms, leads to more powerful representations. Second, we provide direct evidence for our solution to the training
instability problem (Figure~\ref{fig:comprehensive_analysis}(b)). A comparison of training loss curves reveals that our backbone-residual architecture converges significantly faster and to a more stable, lower loss than a naive stacking of heterogeneous experts. This starkly demonstrates that our design effectively mitigates optimization interference, enabling a stable and efficient training paradigm.

\noindent \textbf{Efficiency Analysis}
In addition to performance, we assess the computational efficiency of ADaMoRE against key adaptive baselines. Our analysis is visualized in Figure~\ref{fig:comprehensive_analysis}(c), where bars represent final accuracy and the superimposed line plot represents training efficiency (derived from the inverse of training time on a logarithmic scale). For both metrics, a higher position on the chart indicates a better result. The results reveal that ADaMoRE achieves a compelling balance between high accuracy and computational efficiency. Notably, on the complex Roman-Empire dataset, ADaMoRE not only delivers state-of-the-art predictive accuracy but also exhibits significantly higher training efficiency.

%% file: 5_Related_Works.tex
\vspace{-5pt}
\section{Related Work}
\label{sec:related_work}
\noindent \textbf{Adaptive GNNs on Graphs.}
To handle the structural diversity of graphs, a significant line of research has focused on developing adaptive GNNs. A prominent paradigm within this area is the use of complementary graph filters, often inspired by graph signal processing. These methods typically employ a dual-channel architecture, with a Low-Pass Filter (LPF) to capture smooth, cohesive signals and a High-Pass Filter (HPF) to capture oscillatory, dispersive signals. For instance, GREET~\cite{GREET} and HLCL~\cite{yang2024graphcontrastivelearningheterophily} both learn to distinguish between homophilic and heterophilic edges and apply specialized filters accordingly, although their methods for edge identification and signal fusion differ. Similarly, S3GCL~\cite{S3GCL} utilizes parameterized polynomial filters to create biased (low/high-pass) views for contrastive learning. Other adaptive strategies exist, such as adaptive channel mixing in ACMGNN~\cite{luan2022acm}, which learns to combine feature channels based on their spectral properties. While these approaches demonstrate strong performance, many are tailored to the LPF/HPF dichotomy, which may limit their flexibility. Furthermore, their adaptation mechanisms are often implicitly learned via contrastive objectives, which may not provide a direct or principled signal for fine-grained, node-specific architectural adjustments.

\noindent \textbf{Mixture-of-Experts.}
The Mixture-of-Experts (MoE)~\cite{expert_1,adaptivity, 6796382} paradigm is a more general and powerful framework for adaptivity. The application of MoE to graphs, however, is an emerging field with its own unique challenges.

\textit{Homogeneous vs. Heterogeneous Experts.} Existing Graph MoE approaches can be broadly categorized by their expert composition. A large body of work employs \textbf{homogeneous experts}, where all experts share the same architecture~\cite{li2025hierarchicalmixtureexpertsgeneralizable,wang2025cooperationexpertsfusingheterogeneous,jiang2025adaptivesubstructureawareexpertmodel,ye2025moseunveilingstructuralpatterns}. The primary goal of this approach is often to increase model capacity in a computationally efficient manner, similar to its application in large language models. A more advanced, and arguably more suitable, direction for graph diversity is the use of \textbf{heterogeneous experts}, which combines GNNs with diverse inductive biases~\cite{chen2025mixturedecoupledmessagepassing}. This allows the model to select the most appropriate computational mechanism for a given context.

\textit{Challenges in Unsupervised Graph MoE.} Despite the promise of heterogeneous MoE, training such a system without supervision is notoriously challenging. A core issue is the inherent training instability, that arises from the conflicting optimization objectives of different expert types when guided by a single, weak unsupervised signal. Furthermore, MoE models often suffer from issues like load imbalance and the lack of a clear, explicit optimization objective for the gating network in the absence of labels~\cite{Eigen2013LearningFR,10.5555/3586589.3586709}. While some recent methods have explored heterogeneous experts for graphs, they are either restricted to supervised settings~\cite{hu2021graphclassificationmixturediverse,chen2025mixturedecoupledmessagepassing,ma2024mixturelinkpredictorsgraphs,GMoE} or, when unsupervised, rely on complex expert search mechanisms or customized training strategies~\cite{zhang2024unsupervisedgraphneuralarchitecture}.

%% file: 6_Conclusion.tex
\vspace{-10pt}
\section{Conclusion}
\label{conclusion}
In this paper, we presented ADaMoRE, an unsupervised framework for robust and adaptive graph representation learning, designed to tackle the challenges posed by the diverse and complex structures of modern Web graphs where single GNNs often struggle. Our work addresses the critical gap of training a heterogeneous Mixture-of-Experts framework in a fully unsupervised setting. We introduced a novel backbone-residual architecture that effectively mitigates the training instability inherent in naive expert mixtures. This is achieved by combining a stable foundational backbone with a diverse pool of residually-connected GNN experts, whose functional specialization is explicitly enforced by an information-theoretic diversity regularizer. A structurally-aware gating mechanism performs a fine-grained, node-specific fusion of the learned representations. Trained via a unified unsupervised paradigm, ADaMoRE demonstrates superior generalization and highly competitive performance in both unsupervised and few-shot learning tasks across a range of benchmarks.

%% file: 7.1_Appendix.tex
\appendix

\section{Theoretical Foundations}
\label{ap:theoretical_foundations}

\subsection{Justification for Lemma~\ref{lemma:sample_complexity_reduction} (Sample Complexity Reduction)}
\label{ap:justification_lemma_2}
This section provides a more formal theoretical rationale for Lemma~\ref{lemma:sample_complexity_reduction}. We argue that our backbone-residual architecture is significantly more sample-efficient than a naive MoE, leading to improved training stability. Our analysis is grounded in statistical learning theory.

\noindent \textbf{Theoretical Preliminaries: Sample Complexity and Rademacher Complexity.}
The sample complexity of a learning problem, $N(\epsilon, \delta)$, is the minimum number of samples $n$ required to guarantee that a learned function $\hat{f}$ from a function class $\mathcal{F}$ achieves a true error within $\epsilon$ of the optimal error in that class, with probability at least $1-\delta$. Generalization theory links sample complexity to the complexity of the function class $\mathcal{F}$, measured by the Rademacher complexity, $\mathcal{R}_n(\mathcal{F})$. The required sample size $N$ exhibits a quadratic dependence on this complexity:
\begin{equation}
    N(\epsilon, \delta) \propto \frac{\mathcal{R}_n(\mathcal{F})^2}{\epsilon^2}.
    \label{eq:sample_complexity_rademacher_appendix}
\end{equation}
Thus, reducing the effective Rademacher complexity of the function space is key to improving sample efficiency.

\noindent \textbf{The High Sample Complexity of a Naive Heterogeneous MoE.}
As established in Lemma~\ref{lemma:naive_moe_complexity}, a naive heterogeneous MoE operates within a vast function space, $\mathcal{F}_{Naive-hetero}$. The Rademacher complexity of this space, $\mathcal{R}_n(\mathcal{F}_{Naive-hetero})$, is enormous due to the gated union of disparate function classes. According to Eq.~\ref{eq:sample_complexity_rademacher_appendix}, searching this large space requires a prohibitively high number of samples, manifesting as training instability under a weak unsupervised signal.

\noindent \textbf{The Backbone-Residual Architecture as a Complexity Reduction Strategy.}
Our architecture mitigates this by decomposing the learning problem. We formalize our motivation with the following assumption.

\begin{assumption}[Structured Approximation Capability of the Backbone]
\label{assump:backbone_quality_appendix}
\textit{The oracle function $f^*$ can be decomposed as $f^* = f^*_{struct} + f^*_{res}$, where $f^*_{struct}$ is the "structured" component highly correlated with principal graph structures. As motivated by Proposition~\ref{prop:backbone_efficacy}, our backbone, operating in $\mathcal{F}_{b}$, is designed to effectively capture this component. We assume it can learn a high-quality approximation, $f_{approx} \in \mathcal{F}_b$, such that the norm of the residual target $R^* = f^* - f_{approx}$ is significantly smaller than the norm of the original oracle function:}
\begin{equation}
    \|R^*\| = \|f^* - f_{approx}\| \le \eta \|f^*\| \quad \text{for some constant } \eta \ll 1.
\end{equation}
\end{assumption}

\noindent Given this assumption, the learning task for the residual MoE component is no longer to learn the entire complex function $f^*$, but to learn the much simpler residual function $R^*$. While the full capacity of the residual space $\mathcal{F}_{res}$ might be large, the learning algorithm is guided toward a target with a small norm. This confines the search to a much smaller \textbf{effective subspace}, $\mathcal{F}_{res, eff} = \{f_r \in \mathcal{F}_{res} \mid \|f_r\| \le \eta \|f^*\|\}$.

Standard results in learning theory show that the Rademacher complexity of a function class is proportional to the norm of the functions it contains. Therefore, we can relate the complexities of the effective search spaces:
\begin{equation}
    \mathcal{R}_n(\mathcal{F}_{res, eff}) \propto \eta \|f^*\| \quad \text{and} \quad \mathcal{R}_n(\mathcal{F}_{Naive-hetero}) \propto \|f^*\|.
\end{equation}
Since $\eta \ll 1$, it directly follows that the effective Rademacher complexity of the residual learning task is significantly lower:
\begin{equation}
    \mathcal{R}_n(\mathcal{F}_{res, eff}) \ll \mathcal{R}_n(\mathcal{F}_{Naive-hetero}).
\end{equation}
This leads to the relationship between the sample complexities as stated in Lemma~\ref{lemma:sample_complexity_reduction}. Let $N_{Res}(\epsilon, \delta)$ be the sample complexity for the residual component:
\begin{equation}
    N_{Res}(\epsilon, \delta) \le C \cdot \left( \frac{\mathcal{R}_n(\mathcal{F}_{res, eff})}{\mathcal{R}_n(\mathcal{F}_{Naive-hetero})} \right)^2 N_{Naive}(\epsilon, \delta).
\label{eq:sample_complexity_inequality_appendix}
\end{equation}
Since $\mathcal{R}_n(\mathcal{F}_{res, eff}) \ll \mathcal{R}_n(\mathcal{F}_{Naive-hetero})$, we have $N_{Res}(\epsilon, \delta) \ll N_{Naive}(\epsilon, \delta)$. This significant reduction in sample complexity provides the rigorous theoretical explanation for why our backbone-residual architecture is more stable and efficient. It decomposes the learning task into two more manageable parts: learning the primary structure with the backbone and learning a small-norm correction with the residual experts.

\subsection{Support for Proposition~\ref{prop:backbone_efficacy} (Guided Complementary Filtering)}
\label{ap:support_prop_backbone}
Proposition~\ref{prop:backbone_efficacy} states that the Guided Complementary Filtering paradigm, guided by learned connectivity patterns, serves as a powerful and robust backbone for our architecture. This section provides the theoretical and empirical support for this claim. This principle is grounded in Graph Signal Processing (GSP) and the theory of perfect reconstruction in filter banks.

\noindent \textbf{Theoretical Justification from Perfect Reconstruction.}
From a GSP perspective, node features $\mathbf{X}$ can be viewed as a signal on the graph. A single graph filter, whether low-pass or high-pass, will inevitably lose information by attenuating certain frequency components of this signal. A more robust approach is to use a filter bank with complementary filters to decompose the signal, with the ideal goal of achieving \textbf{perfect reconstruction}---the ability to losslessly recover the original signal from the filtered outputs~\cite{Ekambaram2013GraphSD}.

A classic example is the Simple Spline-like filter bank. For a $d$-regular graph, its Low-Pass Filter (LPF) performs a weighted average, $H_{LP} = \frac{1}{2}(I + \frac{1}{d}A)$, while its High-Pass Filter (HPF) performs a weighted difference, $H_{HP} = \frac{1}{2}(I - \frac{1}{d}A)$. These operators are perfectly complementary, as their sum is the identity matrix, $H_{LP} + H_{HP} = I$, guaranteeing perfect reconstruction. The foundational experts in our backbone, SGC (an LPF) and LapSGC (an HPF), are designed based on this principle of complementarity. Although they do not strictly satisfy the perfect reconstruction property, their core function is to decompose the graph signal into two complementary low- and high-frequency channels, thereby capturing more comprehensive information than any single filter alone.

\begin{table}[h]
\centering
\captionsetup{skip=3pt}
\caption{Performance vs. edge weight numerical distinctiveness.}
\label{tab:ew_numerical_distinctiveness}
\resizebox{0.95\columnwidth}{!}{%
\begin{tabular}{lccccc}
\toprule
Datasets      & 0.9 vs 0.1      & 0.8 vs 0.2      & 0.7 vs 0.3      & 0.6 vs 0.4      & 0.5 vs 0.5      \\ \midrule
Cora          & $\bm{95.03}$ & $92.52$ & $89.01$ & $86.26$ & $83.50$ \\
Citeseer      & $\bm{85.29}$ & $82.37$ & $80.45$ & $75.53$ & $73.53$ \\
Cornell       & $\bm{79.21}$ & $75.53$ & $71.12$ & $73.42$ & $69.47$ \\
Roman\_empire & $\bm{75.21}$ & $75.10$ & $75.06$ & $74.43$ & $73.97$ \\ \bottomrule
\end{tabular}
}
\end{table}

\noindent \textbf{Empirical Investigation of Guiding Weight Quality.} To empirically validate the factors governing the efficacy of our guided filtering backbone, we conducted controlled experiments using oracle edge weights derived from ground-truth labels. The results demonstrate that performance is critically contingent on two key properties of the guiding weights. First, the \textbf{numerical distinctiveness}: a greater numerical separation between the values assigned to cohesive versus dispersive edges consistently yields better performance (Table~\ref{tab:ew_numerical_distinctiveness}). Second, the \textbf{classification accuracy}: the framework is particularly sensitive to the correct identification of the graph's dominant structural pattern, such as cohesive patterns in homophilic graphs (Table~\ref{tab:ew_accuracy_impact}). Collectively, these findings confirm that a clear, accurate, and well-separated signal from the view generator is essential for the filters to perform their specialized functions effectively.

\begin{table}[h]
\centering
\captionsetup{skip=3pt}
\caption{Performance vs. edge weight classification accuracy (cohesive\% / dispersive\%).}
\label{tab:ew_accuracy_impact}
\resizebox{\columnwidth}{!}{%
\begin{tabular}{lcccccc}
\toprule
Dataset & 100/0 & 80/20 & 60/40 & 40/60 & 20/80 & 0/100 \\ \midrule
Cora & \cellcolor{red!70!white}$87.97$ & \cellcolor{red!51!white}$83.24$ & \cellcolor{red!46!white}$82.08$ & \cellcolor{red!37!white}$79.87$ & \cellcolor{red!22!white}$76.35$ & \cellcolor{red!0!white}$70.94$ \\
Computers & \cellcolor{red!70!white}$91.82$ & \cellcolor{red!24!white}$88.71$ & \cellcolor{red!19!white}$88.37$ & \cellcolor{red!18!white}$88.29$ & \cellcolor{red!12!white}$87.81$ & \cellcolor{red!0!white}$86.57$ \\
\midrule
Cornell & \cellcolor{red!0!white}$62.37$ & \cellcolor{red!3!white}$62.89$ & \cellcolor{red!19!white}$65.79$ & \cellcolor{red!25!white}$66.84$ & \cellcolor{red!33!white}$68.42$ & \cellcolor{red!70!white}$75.26$ \\
Texas & \cellcolor{red!0!white}$74.47$ & \cellcolor{red!19!white}$76.84$ & \cellcolor{red!9!white}$75.53$ & \cellcolor{red!0!white}$74.47$ & \cellcolor{red!13!white}$76.05$ & \cellcolor{red!70!white}$83.16$ \\
\bottomrule
\end{tabular}
}
\end{table}

\vspace{-5pt}

\section{Additional Experiments and Analyses}
\label{ap:additional_experiments}

\begin{table*}[t]
\centering
\captionsetup{skip=3pt}
\caption{Performance comparison on the node clustering task (mean \% $\pm$ std. dev.). Best and second-best results are in \textbf{bold} and \underline{underlined}, respectively.}
\label{tab:node_clustering}
\begin{tabular}{l|ccc|ccc|ccc}
\toprule
& \multicolumn{3}{c|}{\textbf{Texas}} & \multicolumn{3}{c|}{\textbf{Actor}} & \multicolumn{3}{c}{\textbf{CiteSeer}} \\
\cmidrule(r){2-4} \cmidrule(lr){5-7} \cmidrule(l){8-10}
Method & ACC & NMI & ARI & ACC & NMI & ARI & ACC & NMI & ARI \\
\midrule
HGRL & $61.97\pm_{3.1}$ & $44.58\pm_{2.1}$ & $37.05\pm_{4.8}$ & $29.79\pm_{1.1}$ & $3.80\pm_{0.8}$ & $4.09\pm_{1.2}$ & $58.06\pm_{4.1}$ & $28.81\pm_{2.0}$ & $21.70\pm_{3.2}$ \\
GREET & $54.32\pm_{6.6}$ & $36.02\pm_{3.6}$ & $23.22\pm_{7.5}$ & $29.28\pm_{0.8}$ & $7.67\pm_{0.5}$ & $3.20\pm_{0.2}$ & $61.14\pm_{1.5}$ & $34.06\pm_{1.7}$ & $33.65\pm_{2.1}$ \\
S3GCL & $57.97\pm_{8.2}$ & $36.83\pm_{5.6}$ & $25.01\pm_{8.5}$ & $26.53\pm_{0.5}$ & $1.59\pm_{0.2}$ & $1.23\pm_{0.4}$ & $65.60\pm_{1.8}$ & \underline{$42.84\pm_{1.6}$} & \underline{$40.45\pm_{3.5}$} \\
\rowcolor{mygray}
ADaMoRE & $\bm{73.58\pm_{3.3}}$ & $\bm{45.08\pm_{4.5}}$ & $\bm{46.86\pm_{6.2}}$ & $\bm{31.21\pm_{0.3}}$ & $\bm{9.19\pm_{0.2}}$ & $\bm{4.18\pm_{0.1}}$ & $\bm{66.56\pm_{2.5}}$ & $40.24\pm_{2.9}$ & $\bm{40.63\pm_{3.8}}$ \\
\bottomrule
\end{tabular}
\end{table*}

\subsection{Empirical Validation of Training Instability}
\label{ap:training_instability}
\begin{figure}[h]
    \centering
    \captionsetup{skip=3pt}
    \includegraphics[width=1\columnwidth]{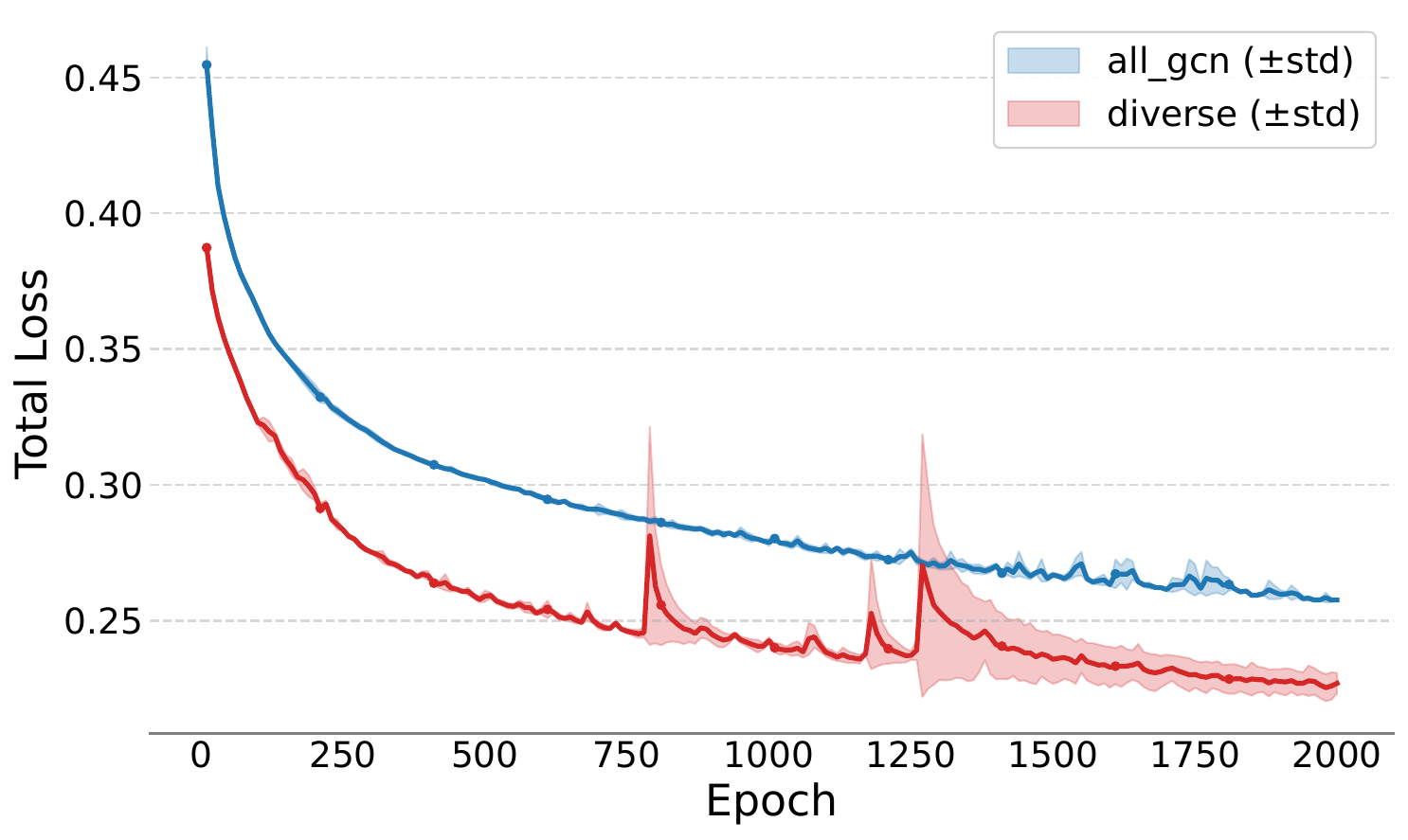} 
    \caption{Training loss curves comparing a homogeneous (all GCN) expert pool against a heterogeneous (diverse GNNs) expert pool within a naive MoE framework.}
    \label{fig:training_instability}
\end{figure}

In our main text, we posit that naively mixing heterogeneous GNN experts leads to inherent \textbf{training instability}. To provide direct empirical evidence for this claim, we conduct a controlled experiment using the GMoE framework~\cite{GMoE} in its original supervised setting. We compare the training dynamics of a homogeneous expert pool (four GCNs) against a heterogeneous one (two GCNs, one GIN, one GraphSAGE) on the Cora dataset. As shown in Figure~\ref{fig:training_instability}, the results are stark: while the homogeneous MoE exhibits a smooth convergence, the heterogeneous MoE suffers from a significantly more volatile and oscillating training process. This experiment empirically validates our core motivation that naively stacking diverse GNNs leads to optimization interference, underscoring the necessity of a principled architectural solution like our proposed backbone-residual design.

\subsection{Evaluation on Node Clustering}
\label{ap:node_clustering}
To further assess the quality of the learned representations beyond classification tasks, we evaluate ADaMoRE's embeddings on a downstream node clustering task. We perform K-Means clustering directly on the pre-trained node embeddings and evaluate the quality using standard metrics: Accuracy (ACC), Normalized Mutual Information (NMI), and Adjusted Rand Index (ARI). As presented in Table~\ref{tab:node_clustering}, ADaMoRE consistently outperforms strong baselines across most datasets and metrics. This demonstrates that the representations learned by our framework possess a more distinct and meaningful cluster structure, further validating their effectiveness in capturing the underlying topology of the graph data.

\subsection{Extended Few-shot Learning Analysis}
\label{ap:extended_few_shot}
\begin{figure}[t]
    \centering
    \captionsetup{skip=3pt}
    
    \begin{subfigure}[b]{0.32\columnwidth}
        \centering
        \includegraphics[width=\linewidth]{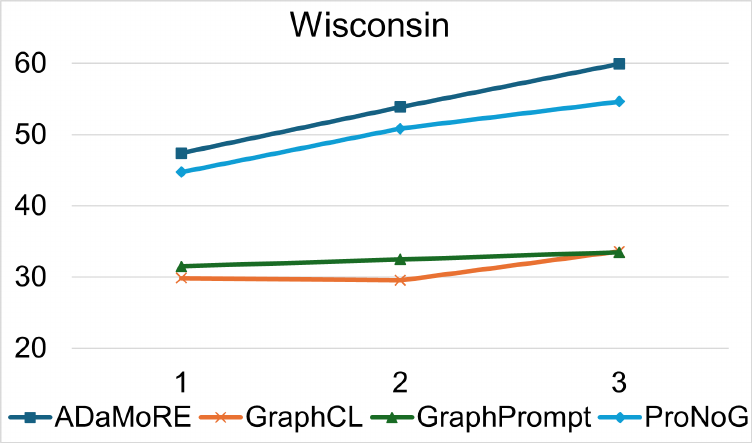}
        \subcaption{Wisconsin}
        \label{fig:fewshot_wisconsin}
    \end{subfigure}
    \hfill 
    \begin{subfigure}[b]{0.32\columnwidth}
        \centering
        \includegraphics[width=\linewidth]{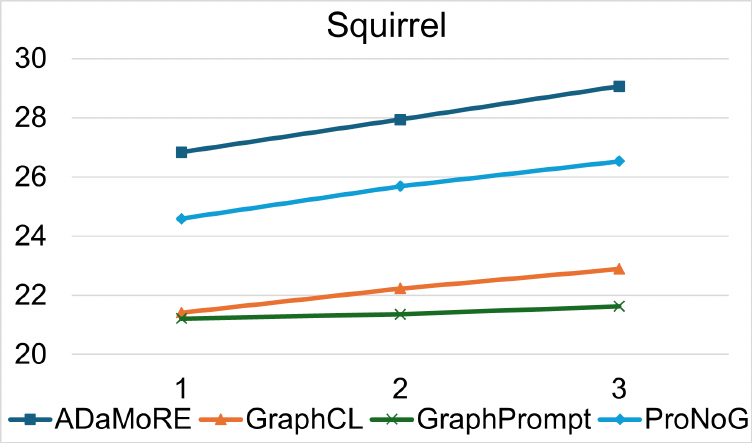}
        \subcaption{Squirrel}
        \label{fig:fewshot_squirrel}
    \end{subfigure}
    \hfill 
    \begin{subfigure}[b]{0.32\columnwidth}
        \centering
        \includegraphics[width=\linewidth]{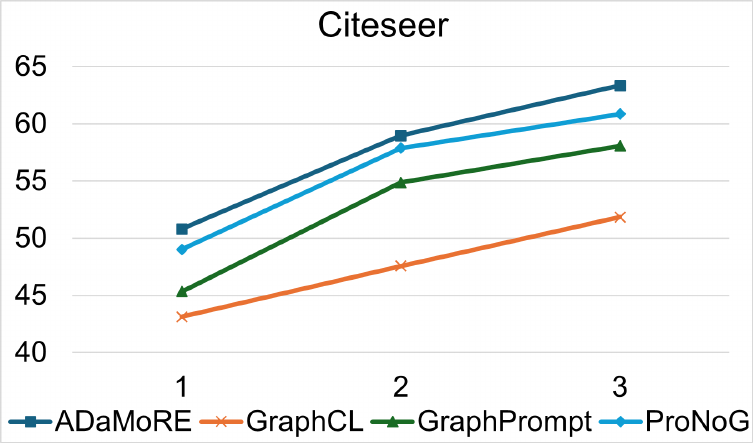}
        \subcaption{Citeseer}
        \label{fig:fewshot_citeseer}
    \end{subfigure}
    
    \caption{Few-shot node classification accuracy of ADaMoRE compared against baseline methods, using 1, 2, and 3 labeled examples per class (shots).}
    \label{fig:few_shot_trends}
    \Description{A composite figure with three line charts for few-shot node classification on the Wisconsin, Squirrel, and Citeseer datasets. Each chart plots accuracy versus the number of shots (1, 2, 3) for ADaMoRE and several other baseline methods. In general, the lines trend upwards as the number of shots increases, and the line representing ADaMoRE is consistently at or near the top.}
\end{figure}

To further assess the robustness of our framework under varying levels of label scarcity, we extended our few-shot node classification experiments to include 1, 2, and 3-shot scenarios. As illustrated in Figure~\ref{fig:few_shot_trends}, ADaMoRE's performance consistently improves with more labeled examples, and more importantly, it maintains a leading or highly competitive performance against baselines across all k-shot settings. This sustained advantage highlights the efficacy of our pre-trained representations for data-efficient learning, demonstrating their capability to adapt effectively even when supervision is minimal.

\subsection{Robustness to Imperfect Guiding Weights}
\label{ap:weight_noise}
To test the framework's robustness, we corrupted oracle-derived edge weights with Gaussian noise at varying ratios. The results, presented in Table~\ref{tab:noise_robustness}, demonstrate a graceful performance degradation as noise levels increase. This confirms the model's resilience to imperfect guiding signals, which we attribute to the interplay between our dual-filter architecture and the adaptive fusion mechanism that can compensate for structural noise.

\begin{table}[h]
\centering
\captionsetup{skip=3pt}
\caption{Performance (Accuracy \%) vs. Noise Level on Guiding Weights.}
\label{tab:noise_robustness}
\resizebox{\columnwidth}{!}{%
\begin{tabular}{lcccc}
\toprule
\textbf{Dataset} & \textbf{No Noise (0.0, 0.0)} & \multicolumn{3}{c}{\textbf{Noise Level (Ratio, StdDev=0.5)}} \\
\cmidrule(lr){3-5}
& & \textbf{(0.2, 0.5)} & \textbf{(0.5, 0.5)} & \textbf{(0.8, 0.5)} \\
\midrule
Cora & 88.96 $\pm$ 1.13 & 88.12 $\pm$ 1.17 & 87.26 $\pm$ 1.82 & 85.74 $\pm$ 1.76 \\
Roman-Empire & 76.59 $\pm$ 0.29 & 75.35 $\pm$ 0.51 & 73.40 $\pm$ 0.53 & 70.44 $\pm$ 0.39 \\
\bottomrule
\end{tabular}
}
\end{table}

\subsection{Hyperparameter Sensitivity Analysis}
\label{ap:hyperparameter_sensitivity}
\begin{figure}[h]
    \centering
    \captionsetup{skip=3pt}
    \begin{subfigure}[t]{0.49\linewidth}
        \centering
        \includegraphics[width=\linewidth]{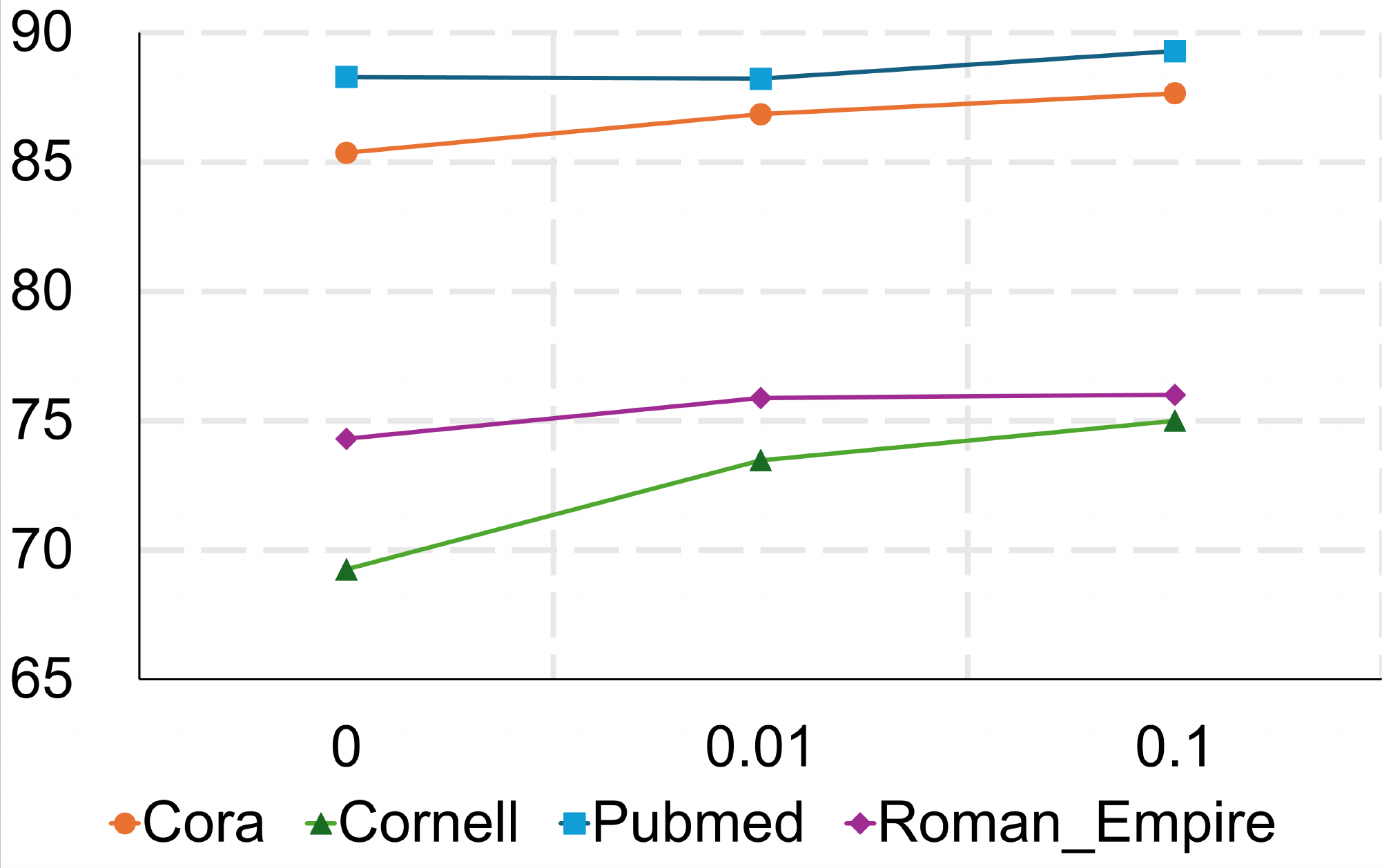}
        \caption{Sensitivity to $\lambda_{load}$.}
        \label{fig:lambda_sensitivity}
    \end{subfigure}
    \hfill
    \begin{subfigure}[t]{0.49\linewidth}
        \centering
        \includegraphics[width=\linewidth]{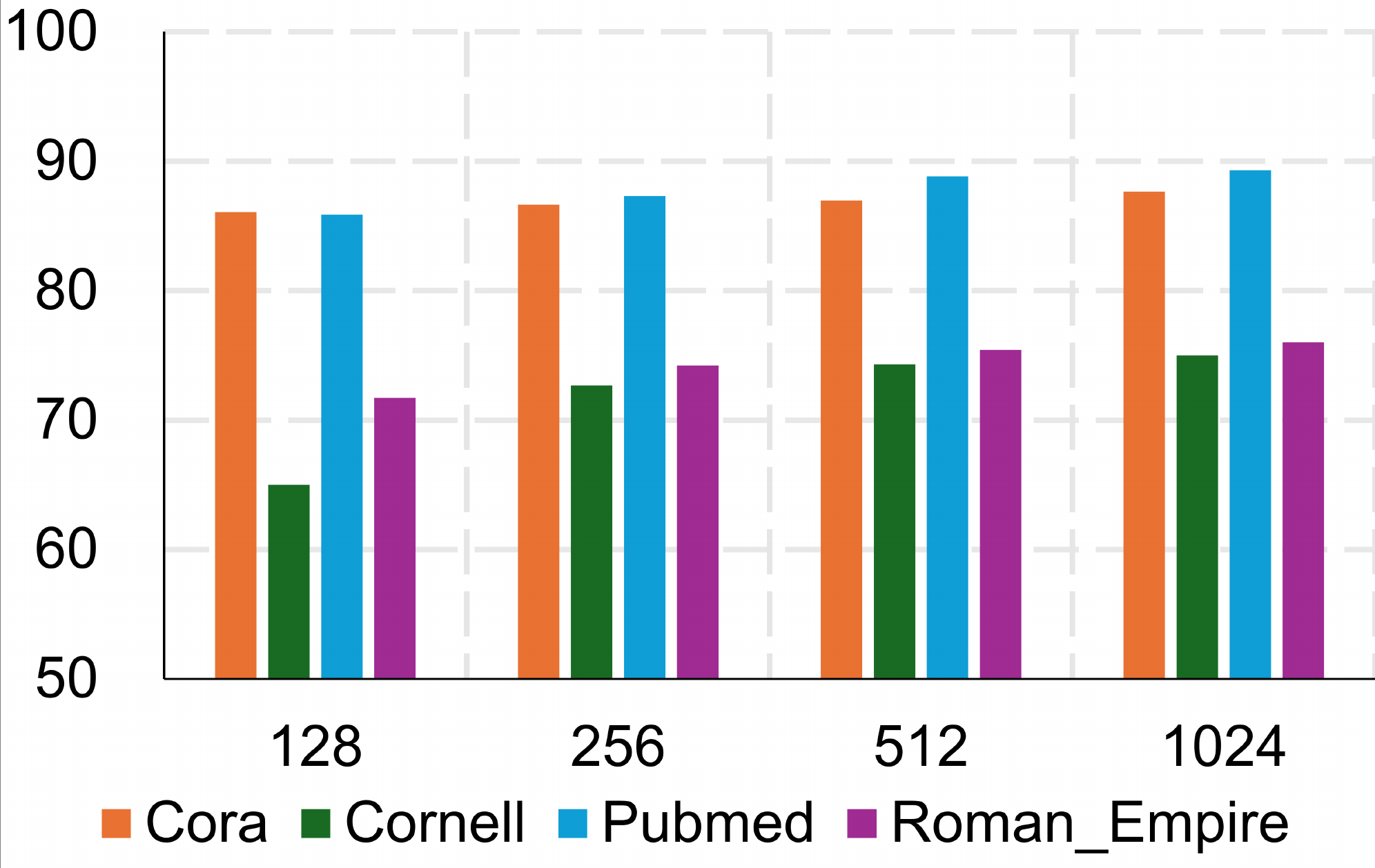}
        \caption{Sensitivity to hidden dimension.}
        \label{fig:hidden_dim_sensitivity}
    \end{subfigure}
    \caption{Hyperparameter sensitivity analysis.}
    \label{fig:sensitivity}
\end{figure}

We analyze the sensitivity of ADaMoRE to the load balancing coefficient, $\lambda_{load}$, and the model's hidden dimension. The results, presented in Figure~\ref{fig:sensitivity}, indicate that our framework is robust within effective operational ranges. As shown in Figure~\ref{fig:sensitivity}(a), a non-zero load balancing coefficient is essential for good performance, with $\lambda_{load} = 0.1$ being optimal in our experiments. Figure~\ref{fig:sensitivity}(b) demonstrates a clear trend where performance progressively improves as the hidden dimension increases from 128 to 1024, suggesting that a larger model capacity is beneficial for learning richer representations.

\section{Implementation Details}
\label{ap:implementation_details}

\subsection{Dataset Details}
\label{ap:dataset_details}
This section provides detailed statistics for the datasets used in our experiments. Table~\ref{tab:un_dataset_statistics} summarizes the benchmarks for the unsupervised node classification task, while Table~\ref{tab:few_shot_dataset_statistics} details those used for the few-shot learning tasks.

\begin{table}[h]
\centering
\captionsetup{skip=3pt}
\caption{Statistics of datasets for unsupervised tasks.}
\label{tab:un_dataset_statistics}
\resizebox{\columnwidth}{!}{%
\begin{tabular}{@{}lccccc@{}}
\toprule
Dataset & Nodes & Edges & Classes & Features & Homophily \\
\midrule
Cora & 2,708 & 10,556 & 7 & 1,433 & 0.81 \\
Pubmed & 19,717 & 88,648 & 3 & 500 & 0.80 \\
Computers & 13,752 & 491,722 & 10 & 767 & 0.78 \\
CS & 18,333 & 163,788 & 15 & 6,805 & 0.81 \\
Ogbn-Arxiv & 169,343 & 1,166,243 & 40 & 128 & 0.66 \\
\midrule
Cornell & 183 & 295 & 5 & 1,703 & 0.31 \\
Texas & 183 & 309 & 5 & 1,703 & 0.11 \\
Wisconsin & 251 & 499 & 5 & 1,703 & 0.20 \\
Roman-Empire & 22,622 & 65,854 & 18 & 300 & 0.05 \\
Minesweeper & 10,000 & 78,804 & 2 & 7 & 0.68 \\
\bottomrule
\end{tabular}%
}
\end{table}

\begin{table}[h]
\centering
\captionsetup{skip=3pt}
\caption{Statistics of datasets for few-shot tasks. (NC: Node Classification, GC: Graph Classification).}
\label{tab:few_shot_dataset_statistics}
\resizebox{\columnwidth}{!}{%
\begin{tabular}{@{}lcccccccc@{}}
\toprule
Dataset & \makecell{Task\\Type} & Graphs & \makecell{Node\\Classes} & \makecell{Graph\\Classes} & \makecell{Node\\Feat.} & \makecell{Avg.\\Nodes} & \makecell{Avg.\\Edges} & \makecell{Hom.\\Ratio} \\
\midrule
Wisconsin & NC & 1 & 5 & - & 1,703 & 251 & 499 & 0.20 \\
Squirrel & NC & 1 & 5 & - & 2,089 & 5,201 & 217,073 & 0.21 \\
Citeseer & NC & 1 & 6 & - & 3,703 & 3,327 & 4,732 & 0.74 \\
\midrule
PROTEINS & GC & 1,113 & 3 & 2 & 1 & 39.06 & 72.82 & 0.66 \\
ENZYMES & GC & 600 & - & 6 & 18 & 32.63 & 62.14 & - \\
BZR & GC & 405 & - & 2 & 3 & 35.75 & 38.36 & - \\
\bottomrule
\end{tabular}%
}
\end{table}

\subsection{Experimental Environment and Model Configurations}
\label{ap:model_configurations}

\noindent\textbf{Experimental Environment.}
All experiments are conducted on a hardware platform equipped with NVIDIA A6000 GPUs and Intel(R) Xeon(R) Platinum 8336C CPU @ 2.30GHz.

\noindent\textbf{Model Configurations.}
For fair comparison, we adhere to standard protocols for unsupervised representation learning~\cite{GREET,S3GCL}. All datasets are sourced from public benchmarks, and for each, we generate ten different random data splits for robust evaluation. Across all models, the hidden dimension is uniformly set to 1024. For all baseline methods, we use their official implementations and follow the hyperparameter settings recommended in their original publications.

Our model, ADaMoRE, is trained with a fixed learning rate of $3\times 10^{-5}$ for 200 epochs using the Adam optimizer. Key components are configured as follows:
\begin{itemize}[nosep, leftmargin=*]
    \item \textbf{Residual Experts:} For the main experiments, to balance efficiency and a fair comparison, we use a single residual expert per channel by default. This expert is selected for each dataset via a hyperparameter search over a candidate pool of \{SGC, LapSGC, GIN, GAT, GraphSAGE, ChebNetII\}. Configurations with multiple residual experts are explored in our ablation studies.
    \item \textbf{Diversity Regularization:} The diversity regularizer $\mathcal{L}_{div}$, is implemented using Centered Kernel Alignment (CKA) with a linear kernel. This loss is applied exclusively among the foundational experts within the sparse MoE backbone to ensure their learned representations are diverse.
\end{itemize}

\noindent\textbf{Few-shot Learning Setup.}
For few-shot learning, ADaMoRE is first pre-trained unsupervisedly. During the downstream adaptation phase, the majority of the expert parameters (both backbone and residual) are kept frozen. Only the parameters of the adaptive components, namely the Structurally-Aware Gating module and the final Adaptive Fusion Gating, are fine-tuned on the few labeled instances.